\documentclass{article}
\usepackage{frExamplee}
\usepackage{graphicx}
\usepackage{apalike}
\usepackage{setspace}

\usepackage{float}
\usepackage{cite} 
\usepackage{amsmath}
\usepackage{amssymb}
\usepackage{graphics} 
\usepackage{epsfig} 
\usepackage{balance}
\usepackage{xcolor}

\title{Scale-Invariant Specifications for \\Human-Swarm Systems}

\author{
Joel Meyer \thanks{Authors contributed equally to this work} \thanks{Authors are affiliated with the Center for Robotics and Biosystems at Northwestern University} \\
\And
Ahalya Prabhakar \footnotemark[1] \thanks{Author is affiliated with École Polytechnique Fédérale de Lausanne (EPFL)}\\
\AND
Allison Pinosky \footnotemark[2]\\
\And
Ian Abraham  \thanks{Author is affiliated with the Department of Mechanical Engineering at Yale University}\\
\And
Annalisa Taylor \footnotemark[2]\\
\And 
Millicent Schlafly \footnotemark[2]\\
\And 
Katarina Popovic \footnotemark[2]\\
\And 
Giovani Diniz \thanks{Author is affiliated with Unmanned Laboratory}\\
\And 
Brendan Teich
\thanks{Authors are affiliated with Raytheon BBN Technologies}\\
\And 
Borislava Simidchieva \footnotemark[6]\\
\And 
Shane Clark \thanks{Author is affiliated with Systems and Technology Research}\\
\And 
Todd Murphey \footnotemark[2]\\
}

\begin{document}

\maketitle

\begin{abstract}
We present a method for controlling a swarm using its spectral decomposition---that is, by describing the set of trajectories of a swarm in terms of a spatial distribution throughout the operational domain---guaranteeing scale invariance with respect to the number of agents both for computation and for the operator tasked with controlling the swarm.  We use ergodic control, decentralized across the network, for implementation.  In the DARPA OFFSET program field setting, we test this interface design for the operator using the STOMP interface---the same interface used by Raytheon BBN throughout the duration of the OFFSET program.  In these tests, we demonstrate that our approach is scale-invariant---the user specification does not depend on the number of agents; it is persistent---the specification remains active until the user specifies a new command; and it is real-time---the user can interact with and interrupt the swarm at any time.
Moreover, we show that the spectral/ergodic specification of swarm behavior degrades gracefully as the number of agents goes down, enabling the operator to maintain the same approach as agents become disabled or are added to the network. We demonstrate the scale-invariance and dynamic response of our system in a field relevant simulator on a variety of tactical scenarios with up to 50 agents. We also demonstrate the dynamic response of our system in the field with a smaller team of agents. Lastly, we make the code for our system available.
\end{abstract}

\section{Introduction}
One of the biggest problems in the field of swarm robotics is that it is difficult for human users to reap the potential benefits of swarm robotic systems (i.e., robustness, information gains) due to the high cognitive load associated with controlling swarms \cite{durantin_using_2014}. Cognitive load has been shown to scale with both the size of the swarm a user controls, as well as with the complexity of the environment in which the swarm and user operate (due to both agent-agent interactions and the length of time the operator has to reason about decisions) \cite{gateau_considering_2016}. Swarm operators need interfaces and control algorithms that are \emph{scale-invariant} to enable operators to specify the same objective to swarms of different sizes without modification. Two major components are needed to create such a system---an interface to transform user input into scale-invariant commands and a control algorithm to automatically and flexibly adapt user commands to changes in swarm size (which may occur due to communication dropouts, hardware failures, or newly available robots joining the swarm). 

\subsection{Scale-Invariant User Interfaces}
Prior work on user interfaces for swarm systems include interfaces that enabled operators to control swarms of various sizes through haptic devices \cite{lee_haptic_2011, tsykunov_swarmtouch_2019}, touch-based tablet applications \cite{ayanian_controlling_2014, kato_multi-touch_2009}, and whole-body spatial gestures \cite{podevijn_gesturing_2014, nagi_human-swarm_2014}. Other researchers have run studies to find sets of gestures users found natural for controlling a swarm \cite{kim_user-defined_2020, micire_analysis_2009} or developed programming languages that made it easier for users to control a robot swarm through a computer terminal \cite{pinciroli_buzz_2016}. While these methods enabled a direct mapping between human commands and swarm motion, they did not provide a method for controlling the swarm's behavior beyond the initial specification. Operators had to maintain constant situational awareness to input new commands for their swarm. 
Prior implementations of swarm interfaces in field scenarios only displayed information and did not allow operators to modify the behavior of their swarm through the interface \cite{mccammon_ocean_2021}, or limited users by only allowing them to modify their swarm's behavior through trajectory waypoints \cite{leonard_coordinated_2010}. Other swarm interfaces for the field limited users by only allowing them to select pre-defined behaviors from a ``playbook'' \cite{hsieh_adaptive_2007}.

As a swarm interacts with both the environment and other agents in the environment, the operator's interface should enable the operator to specify behavior that is both scale-invariant with respect to swarm size and \emph{persistent}---which gives the operator the option of performing other tasks while the swarm executes their initial command. The interface should also enable users to modify these scale-invariant and persistent commands while they are in-progress if the environment or user's task change. The user swarm interface we extend in this paper, initially introduced in \cite{prabhakar_ergodic_2020}, enables users to specify commands to their swarm as target distributions that can be deployed regardless of swarm size. The user can modify these commands in real-time, and attend to other tasks after they specify a command since the swarm will persistently converge to the target given by the user. User target distributions are not pre-defined. Users can draw their own target distributions, which may correspond to another user's high-level conception of how to execute a task even if the two operators' target distribution drawings are qualitatively different.

\subsection{Scale-Invariant Control Algorithms}
Previous work in swarm control algorithms has used methods that pre-specify motion behaviors for each robot in the swarm as part of a formation for accomplishing a task \cite{giles_mission-based_2017, balch_behavior-based_1998, setter_team-level_2015, kolling_towards_2012, bevacqua_mixed-initiative_2015}. Other previous work designed specifically for swarm sensing and coverage has relied on potential fields \cite{ogren_cooperative_2004, song_potential_2002, baxter_multi-robot_2007} and/or Voronoi partitions \cite{cortes_coverage_2004, schwager_decentralized_2009, pimenta_sensing_2008} to plan swarm trajectories. While these methods have also been used for applications such as target tracking \cite{lee_tracking_2010} and swarm agent rendezvous \cite{cortes_robust_2006}, they were not designed for dynamic re-planning with a human user or for re-planning needs due to changes in swarm size from attrition or newly available agents joining the swarm. More recent swarm control methods have exhibited a greater degree of flexibility in adapting to changes in swarm size but are controlled via a central node, which leads to swarm system vulnerability \cite{sampedro_flexible_2016, wei_agent-based_2013}. Other work on exploration and mapping \cite{alonso-mora_distributed_2019, schwager_decentralized_2009} incorporates decentralized methods that are robust to swarm agent communication dropouts and hardware failure. However, these methods do not enable a user to enter the swarm control loop and flexibly re-specify swarm behavior as the swarm's environment and or user objectives change.

Various attempts have been made to enable human users to flexibly re-specify swarm behavior. In ~\cite{swamy_scaled_2020}, a user can take control of one member of their swarm to influence or modify the task the swarm as a whole is currently performing. This work bears similarities to other prior work involving leader-follower methods \cite{walker_human_2013, goodrich_leadership_2012} for controlling and influencing a swarm's behavior with individual swarm members or small sub-teams within the swarm. However, with an increasing number of agents, adjusting swarm behaviour by influencing individual agents can become less effective if the user's swarm can lose designated leader agents to attrition (communication dropouts, hardware failure, etc.). 

In contrast, defining how the collective swarm behaves using a spatial distribution has been shown to both scale favorably to swarms of different sizes, and enable human users to dynamically re-plan with swarms. Spatial distributions (both static and dynamic) have been used to control swarms in sensing and coverage applications \cite{smith_persistent_2012, smith_persistent_2011, ghaffarkhah_dynamic_2011}. Dynamic distributions created by a human user have especially shown promise for real-time human swarm control. The authors of \cite{diaz-mercado_distributed_2015} present a decentralized, density-based coverage approach in which an operator uses a tablet interface to define an area for their swarm to explore. The user draws the desired exploration areas on the tablet interface, which are then partitioned into different regions using Voronoi tesselations. These tesselations are assigned to individual agents in the swarm for coverage. Each agent then needs to communicate with its nearest neighbors for the swarm to converge. This system as a whole was robust to agent dropouts due to decentralization, was demonstrated with swarms of different sizes, and enabled users to dynamically re-plan by redrawing the desired swarm coverage distribution on their tablet.

The method we extend from \cite{prabhakar_ergodic_2020} uses a decentralized ergodic coverage algorithm (first introduced in \cite{abraham_decentralized_2018}), which contains many of the desirable attributes from \cite{diaz-mercado_distributed_2015} but does not need to partition the operating space and assign partitions to each agent in the swarm. A robot running decentralized ergodic coverage spends an amount of time in different regions of the exploration space that is proportional to the spatial statistics measure of these regions in the exploration space, taking into account the consensus estimate of how much time the entire swarm has spent there. This implies a human operator can use spatial distributions to specify how a swarm should allocate its agents to achieve proportional coverage of an exploration space in a decentralized manner. Unlike the algorithm presented in \cite{diaz-mercado_distributed_2015}, swarms of agents using decentralized ergodic coverage  do not need to communicate with their nearest neighbors to converge to a user's distribution---communication with any agents in the swarm will improve convergence \cite{abraham_decentralized_2018}. Not using Voronoi partitions also enables individual agents in the swarm to explore the full environment without being bound to a particular assigned area. This is especially helpful under agent communication dropout or hardware failure since there is no need to recalculate or reassign partitions---our system will persistently and flexibly adapt to changes in swarm size and the user target distribution.

\subsection{Main Contributions}
Our work develops a method for scale-invariant swarm control with a touch interface and a decentralized control algorithm. The touch interface enables users to prescribe strategies at the swarm-level rather than at the individual agent-level \cite{prabhakar_ergodic_2020}. Gestures from the touch interface are converted to commands for the swarm using a decentralized ergodic coverage approach, which is invariant to swarm size \cite{abraham_decentralized_2018}. Each agent responds to the desired coverage map in real-time. We extend our prior work described in \cite{prabhakar_ergodic_2020} by demonstrating the capability of our system with larger swarms with field-relevant executions in the STOMP (Swarm Tactics and Offensive Mission Planning) command software.

Our main contributions are as follows: \textbf{1)} user interfaces for creating scale-invariant specifications for human-swarm systems; \textbf{2)} a scale-invariant (with respect to number of agents) integrated system architecture for
real-time human swarm control that enables dynamic re-planning with swarms. These two components enable human operators to dynamically re-plan with a swarm for exploration and distributed sensing by using ergodicity as a quantitative measure of information in the environment. Using our system, the operator maps visual input from both their line of sight in the environment and from the command interfaces at their disposal (described in Section \ref{sec:STOMP}) to a spatial representation of information that they then send as a command to their swarm via their user interface. Each member of the swarm runs a decentralized ergodic coverage algorithm (described in Section \ref{sec:algorithms}) that transforms user commands into swarm trajectories. The combination of visual input the operator receives, the user interfaces they send swarm commands through, and the decentralized ergodic coverage algorithm running on each member of the swarm enables the swarm to converge to the operator's spatial representation of information regardless of how many agents in the swarm the operator has at their disposal at any point in time. Since the internal process an operator uses to map visual input to expected information content is qualitative (in the sense that we do not have a model of how humans make these choices), our system affords flexibility to the operators---different operators can send different commands (representing spatial information) to their swarms based upon the same visual input.  

\section{System Overview}\vspace{-0.2em}
In this section, we introduce the key elements of our system. First, we describe the user interfaces. Then we describe the mission planning interface. Then, we discuss the individual swarm agents (rovers). Finally, we discuss the overall architecture linking these components. 

\subsection{The User Interface}
\label{sec:tanvas_and_touchscreen}
\begin{figure}[htb]
    \centering
    \includegraphics[width=0.8\textwidth]{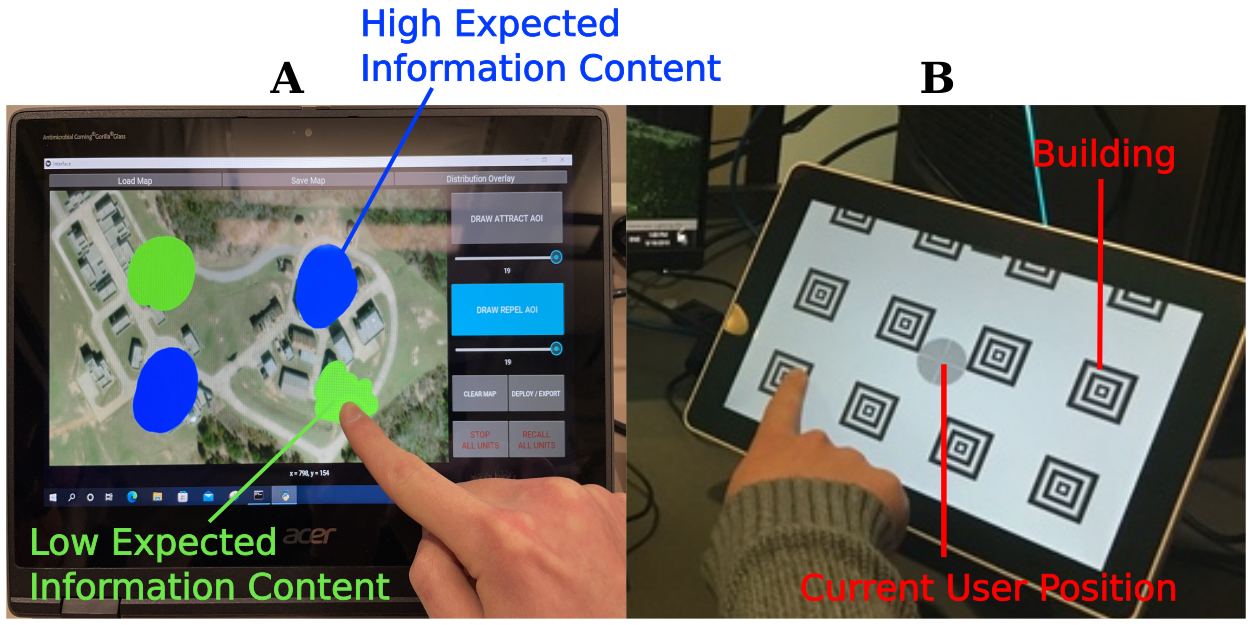}
    \caption{\textbf{User Specifications Using Touch Interfaces:} Figure \ref{fig:combined_touchscreen_tanvas}A shows the touchscreen interface, through which users can draw areas containing high (blue) and low (green) expected information content with their fingertips. Users draw over a map of the swarm's operating environment and can adjust their drawings in real-time. Figure \ref{fig:combined_touchscreen_tanvas}B shows the Tanvas interface, which renders information in the environment as changes in surface friction the user feels as they slide their finger across the interface, which enables the user to localize themselves in the environment and issue commands without looking at the interface. The circle on the Tanvas interface represents the user's current position in the environment, while the squares represent buildings in the environment. Users can also draw high and low expected information content areas with the Tanvas.  We used the touchscreen for our field relevant simulations in  Section~\ref{sec:stomp_simulations} and the Tanvas for field tests in Section~\ref{sec:fieldexercises}.}
    \label{fig:combined_touchscreen_tanvas}
\end{figure}

Over the course of the DARPA OFFSET program we developed two user interfaces. A touchscreen interface was used for experiments run with Raytheon BBN's STOMP command software---where the only difference between simulation and physical experiments is that STOMP is communicating with simulated agents. Additionally, a Tanvas interface \cite{olley_electronic_2020} with haptic feedback was used for physical hardware experiments at OFFSET field exercises. Both interfaces are shown in Figure \ref{fig:combined_touchscreen_tanvas}. Details of both interfaces are included in the sections below.

\subsubsection{Touchscreen}
\label{sec:touchscreen}
The touchscreen interface shown in Figure \ref{fig:combined_touchscreen_tanvas}A transmits user specifications to the user's swarm of agents. \textbf{``User specifications''} (aka \textbf{``target distributions''}) are distributions in physical space that convey areas the user wants their swarm to converge to or avoid for the purpose of exploration. The operator is able to draw these areas on the touchscreen with their fingertip, stylus, or computer mouse. Blue regions indicate exploratory areas (high expected information content) for agents, and green regions specify restricted areas (low expected information content) that the agents must avoid. \textbf{``Expected Information Content''} in this work is defined as the relative importance an operator assigns to a particular area of the environment (via prior knowledge or intuition) for the purpose of exploration with their swarm. Areas with high expected information content may be expected to contain a high-value target, while areas with low expected information content may be expected to not contain a high-value target (or may be suspected to contain dangers that could disable the user's swarm). 

The user can adjust the thickness of the lines they draw while creating these regions (for finer or coarser specifications) by using the slider interface on the right side of the touchscreen. User specifications that contain more than one target area (i.e., both a blue and a green region, multiple blue regions, multiple green regions, or a combination of the aforementioned) are \textbf{``multimodal''}, and represent a multimodal distribution of expected information in physical space. Once the user finishes drawing these regions they can tap/click the “Deploy” button, which sends the user's multimodal drawing (in 2-D array form) to each agent in the swarm through a ROS websocket. 
Additionally, the touchscreen interface allows the user to generate an RGB image of the map with the user's multimodal distribution overlaid on top of it for debugging purposes post-experiment or for situational awareness during testing to see how the user's drawing is translated into a multimodal distribution. User specifications are modifiable in real-time; users can draw over regions that have already been drawn on, even if the agents have not yet made it to those regions. Users can also clear the interface screen if they wish to redo a particular target specification. 

The touchscreen interface application can be loaded on either tablets or PC's. Users can launch multiple instances of the interface application if the user wants to specify target distributions for different subgroups of the swarm (keeping two or more specification channels for these sub-teams open at the same time). The ability to use the touchscreen interface application on either tablets or PC's gives the operator flexibility in not being restricted to a particular operating system. A downside of the touchscreen interface is that there is no haptic feedback. The operator needs to continuously look at the screen while they specify target distributions, meaning the operator may struggle to re-specify swarm behavior if their operating environment requires visual awareness.

\subsubsection{Tanvas}\label{sec:tanvas}
The Tanvas (shown in Figure \ref{fig:combined_touchscreen_tanvas}B), on the other hand, gives users haptic feedback reflecting real-time changes in the environment as they send commands to their swarm. The Tanvas renders textures on its smooth screen by modulating the friction underneath the user's fingertip \cite{olley_electronic_2020}. These haptic features represent different objects in the environment, such as buildings, with different textures. The circle on the Tanvas interface screen represents the user's current position in the environment, whereas squares on the Tanvas interface screen represent buildings and other structures. The Tanvas renders all features on its screen with respect to the user's body frame. 
With the Tanvas, the user can specify regions of exploratory interest by double-tapping the screen, shading the region on the Tanvas, and double-tapping the screen again. At the second double-tap, the user's specification coordinates are transformed into a distribution and then sent over a TCP socket to each agent in their swarm. The Tanvas also provides audio feedback to the user when they double-tap the screen.

Haptic feedback is critical for ``boots on the ground''. Operators need to be aware of their immediate surroundings while working and cannot afford to keep their eyes focused on an interface. The Tanvas enables users to orient themselves in the environment and send commands while watching what is going on around them without looking away, which in the OFFSET scenario was particularly useful for scenarios that required quick re-specification of the swarm and that also required high-operator awareness of their surroundings due to drones flying close by. A downside of the Tanvas is that it represents the operating environment at a lower level of granularity with haptic feedback than the touchscreen interface can with visual feedback.

\subsection{STOMP}
\label{sec:STOMP}
\begin{figure}[!htb]
    \centering
    \includegraphics[width=0.8\textwidth]{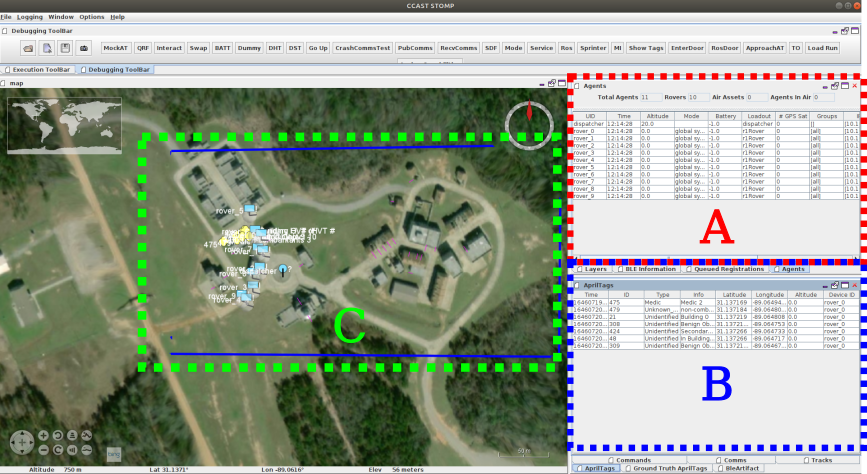}
    \caption{\textbf{STOMP -- A Field Relevant Simulator for OFFSET:} This figure shows an overview of the STOMP simulator. Users monitor the positions of their swarm and receive real-time information about the status of each agent, communications between agents, and information detected by agents as they operate in the environment. Boxes A and B show panels that display agent status, communications, and April Tag detections. Box C shows the agents' real-time positions in the operating environment.}
    \label{fig:STOMP}
    \vspace{-1em}
\end{figure}

As part of OFFSET, we fully integrated our decentralized ergodic coverage algorithm with Raytheon BBN's mission planning interface  STOMP (Swarm Tactics and Offensive Mission Planning). Figure \ref{fig:STOMP} shows an overview of the STOMP environment. Box A shows the status of each rover, which includes its IP address and battery level. Box B displays information about different objects detected by the rovers. Since the objects are identified by April Tags, box B contains the ID number of each tag, the type of information the tag represents (medic, civilian, building, etc.), the global pose of the tag in the environment, and the ID number of the rover that detected it. Box C shows the current state of the operating environment, which includes real-time rover positions and detected information.

\subsection{Rover}
Figure \ref{fig:unifyingarchitecture} shows the Aion Robotics R1 Rover that was used for field tests performed at DARPA OFFSET. The rover includes an NVIDIA Jetson TX-2 embedded computing device and a Pixhawk 2.1 controller with a HERE GPS unit running ArduRover 3.2. Onboard sensors include an RPLidar A1M8 2D 360 Lidar and an Intel RealSense D435i depth camera. 
Motion planning on the rover used the decentralized ergodic coverage algorithm, described below in section \ref{sec:algorithms}, whose output was sent to an RT-RRT$^\star$ (Real-Time Rapidly Exploring Random Tree) algorithm \cite{naderi_rrt}, described in section \ref{sec:rrt}, for real time path planning and obstacle avoidance. 

The ergodic coverage algorithm ran locally on the NVIDIA TX-2 of each individual rover running a local ROS network. The rovers in the swarm communicated with each other and received swarm-level commands over a local LTE network through a Java/Protelis interface developed by the Raytheon BBN team. The rovers connected to the LTE basestation via custom-built USB LTE modems.
The OFFSET experiments were conducted on a small range area that contained a combination of grassy terrain and concrete sidewalks. The rovers operated in a bounding box defined by GPS coordinates provided by Raytheon BBN's Java/Protelis interface. 
User commands were given at the Tanvas interface, which was connected to the robotic agents through the Java/Protelis interface and a TCP protocol. 

\subsection{Unifying System Architecture}
\label{system_architectures}
\begin{figure}[htb]
    \centering
    \includegraphics[width=1.0\textwidth]{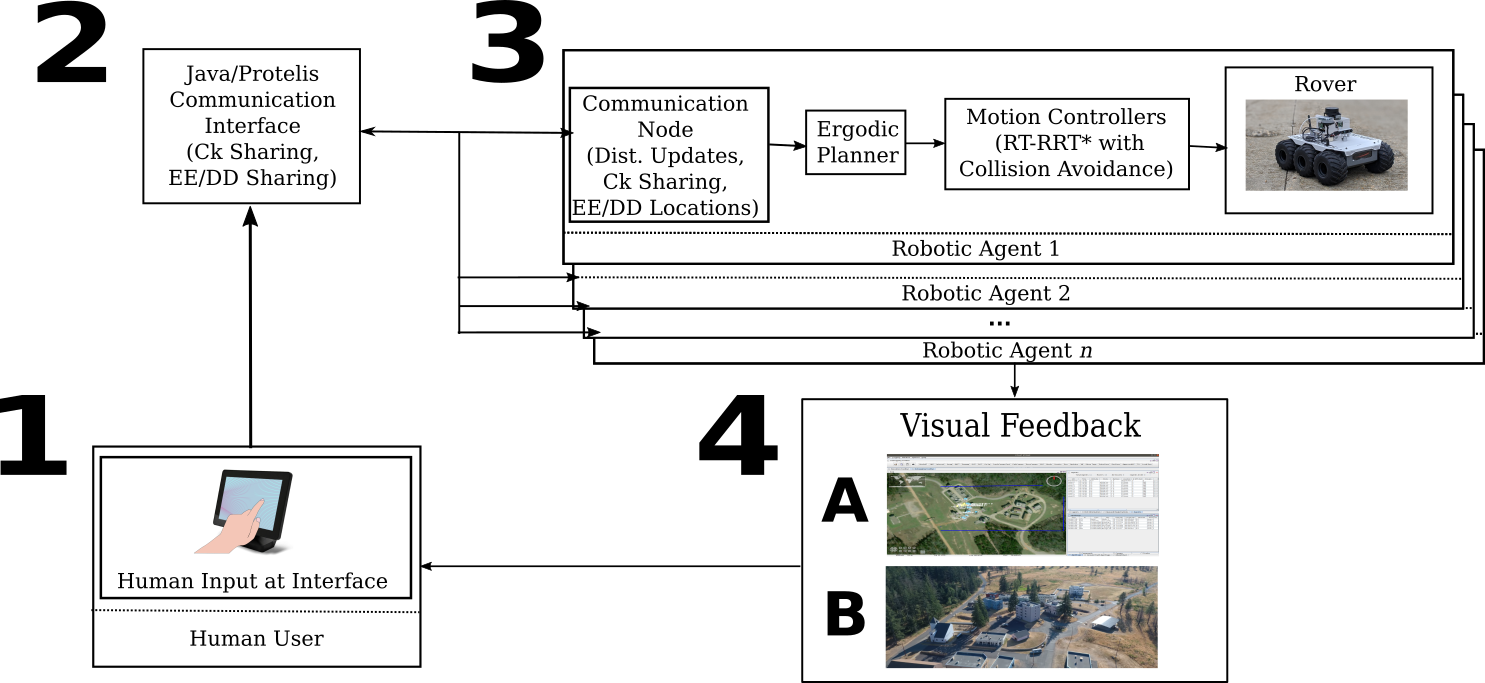}
    \caption{\textbf{A Unifying System Architecture:} This figure shows the entire sequence of human-swarm interaction in our system. In (1) the user starts by inputting their target distribution through an interface (either a Tanvas or touchscreen). The target distribution is sent to a communication node running on each rover as shown in (2). The input is then transformed into a set of control commands via the algorithm running on each rover (explained in Section \ref{sec:algorithms}) in (3). The operator uses visual feedback on the swarm's progress in (4)---provided through plots (introduced in Section \ref{sec:stomp_simulations}), the locations of the agents in STOMP (4A), and the operator's line of sight in the actual environment (4B)---to plan subsequent target distributions to send to the rovers to accomplish their desired task.}
    \label{fig:unifyingarchitecture}
\end{figure}
Figure \ref{fig:unifyingarchitecture} shows a unifying system architecture that contains the major components of our system that were used throughout the OFFSET program. It also shows the entire sequence of human-swarm interaction for a user commanding a swarm with our system. Users start by inputting commands for their swarm through gestures at their user interface (either a Tanvas or touchscreen) that are transformed into target distributions representing swarm-level behavior and then sent to a communication node running on each rover. The algorithm running on each rover (explained in Section \ref{sec:algorithms}) uses the target distribution it has received via its communication node to determine its next control actions. Rovers send their past trajectories to each other (described in section \ref{sec:algorithms}), which enables the whole team of rovers to converge to the user's commanded target distribution. The user receives visual feedback on the progress of their swarm as it converges to their target distribution via the locations of the agents in their swarm through STOMP and the user's line of sight in the actual environment. The user also receives feedback on the progress of their swarm via plots (introduced in Section \ref{sec:stomp_simulations}). All of this information will help the user determine the content and timing of their next commands/target distribution specifications as they complete their desired task.

\section{Algorithm Overview}\label{sec:algorithms}

This section describes a decentralized ergodic coverage algorithm (first derived in \cite{abraham_decentralized_2018}) that converts spatially distributed task information into temporally driven motion for a networked set of robotic agents. To integrate this algorithm with our system, we have extended the algorithm by augmenting its outputted agent trajectories with RT-RRT$\star$ for low-level obstacle avoidance. The inputs to the algorithm are high-level user task specifications that are distributions in physical space representing expected information content in the environment. Incorporating user task specifications into the decentralized ergodic coverage algorithm enables dynamic re-planning with the user's swarm for exploration tasks.

    \subsection{Ergodicity and the Ergodic Metric}

        We start by defining ergodicity and the ergodic metric. Assume the state of a single robotic agent at time $t$ is given by $x(t) : \mathbb{R}^+ \to \mathbb{R}^n$ and the controls given to the robot at time $t$ are defined as $u(t) : \mathbb{R}^+ \to \mathbb{R}^m$. The dynamics of the robot are defined to be a control-affine dynamical system of the form

        \begin{equation} \label{eq:robot_dynamics}
            \dot{x}(t) = f(x(t),u(t)) = g(x(t)) + h(x(t)) u(t),
        \end{equation}
        
        where $g(x) : \mathbb{R}^n \to \mathbb{R}^n$ is the free, unactuated dynamics of the robot and $h(x):
        \mathbb{R}^n \to \mathbb{R}^{n \times m}$ is the dynamic control response subject to input $u(t)$. Define the robot's time-averaged statistics $c(s, x(t))$ for a trajectory $x(t)$ (i.e., the statistics describing
        where the robot spends most of its time) for some time interval $t \in \left[ t_i, t_i + T\right]$ as
        
        \begin{equation}\label{eq:time_avg_stats}
            c(s, x(t)) = \frac{1}{T}\int_{t_i}^{t_i+T} \delta (s - x_v(t)) dt,
        \end{equation}
        
        where $\delta$ is a Dirac delta function, $T \in \mathbb{R}^+$ is the time horizon, $t_i \in \mathbb{R}^+$ is
        the $i^\text{th}$ sampling time, and $x_v(t) \in \mathbb{R}^v$ is the state that intersects with the exploration space (with state dimension $v <$ less than dimension $n$ of the exploration space). The ergodic metric introduced in \cite{mathew_metrics_2011} relates the time-averaged statistics
        $c(s,x(t))$ to an arbitrary spatial distribution $\phi(s)$ via:
        
        \begin{align} \label{eq:ergodic_metric}
            \mathcal{E}(x(t)) & = q \,\sum_{k \in \mathbb{N}^v} \Lambda_k \left(c_k -\phi_k \right)^2   \\
            & = q \, \sum_{k \in \mathbb{N}^v} \left( \frac{1}{T} \int_{t_i}^{t_i + T} F_k(x(t)) dt - \phi_k \right)^2 \nonumber,
        \end{align}
        where
        \begin{equation*}
            \phi_k =  \int_{\mathcal{X}_v} \phi(s) F_k(s) ds.
        \end{equation*}
        In this work, $\phi(s)$ is the operator's spatial representation of information in the environment based upon their line of sight in the environment and visual input they receive from their command interfaces (Section \ref{sec:STOMP})---in other work such as \cite{miller_ergodic_2016}, $\phi(s)$ is constructed from an agent's sensor's measurement model. $q \in \mathbb{R}^+$ is a scalar weight on the metric, $c_k$ is the Fourier decomposition of $c(s,x(t))$~\footnote{The
        cosine basis function is used; however, any choice of basis function $F_k$ that can be differentiated with respect to the state $x(t)$ and that can be evaluated along an agent's trajectory is acceptable.}, $\phi_k$ is the \emph{spectral decomposition} of the spatial distribution $\phi(s)$, and
        \begin{equation*}
            F_k(x) = \frac{1}{h_k}\prod_{i=1}^v \cos \left( \frac{k_i \pi x_i}{L_i} \right)
        \end{equation*}
        is the cosine basis function for a given coefficient $k \in \mathbb{N}^v$. $h_k$ is a normalization factor
        defined in ~\cite{mathew_metrics_2011} and $\Lambda_k = (1 + \Vert k \Vert^2)^{-\frac{v+1}{2}}$ are weights on the
        frequency coefficients. A robot whose trajectory $x(t)$ minimizes (\ref{eq:ergodic_metric}) as $t\to \infty$ is
        optimally ergodic with respect to the target distribution. That is, the robot spends an amount of time in
        different regions of the exploration space that is proportional to the spatial statistics measure of these regions in the exploration space.

       Thus, we obtain a method for specifying how
        long a single agent should spend in different regions of the exploration space. If the user sends a multimodal target distribution to the system that conveys the relative ``importance'' of each area, minimizing the ergodic metric enables an agent to spend time in each area that is proportional to each area's expected information content. If the user needs to modify their target distribution, the agent will flexibly respond by changing its trajectory to match the new distribution of expected information content in the environment. The next section shows that in addition to allowing flexible task specification, the ergodic metric enables a user to control a decentralized network of agents.


    \subsection{Decentralized Ergodic Control} \label{subsec:decentralized-ergodic-control-using-consensus}
        Consider a set of $N$ agents with state $x(t) = \left[ x_1(t)^\top, x_2(t)^\top, \ldots, x_N(t)^\top\right]^\top :
        \mathbb{R}^+ \to \mathbb{R}^{n N}$. The multi-agent system's contribution to the time-averaged statistics $c_k$ can be rewritten as
        \begin{align}\label{eq:centralized_ck}
            c_k & = \frac{1}{N} \sum_{j=1}^N
                    \frac{1}{T_\mathcal{E}} \int_{t_i}^{t_i + T} F_k(x_j(t)) dt \nonumber \\
            & = \frac{1}{T_\mathcal{E}} \int_{t_i}^{t_i+T} \tilde{F}_k(x(t)) dt,
        \end{align}
        where $\tilde{F}_k(x(t)) = \frac{1}{N}\sum_j F_k(x_j(t))$, $N$ is the number of agents, $T$ is the time horizon, $t_i$ is the current time step, $\Delta t_\mathcal{E}$ determines how far into the past the agent needs to remember, and $T_\mathcal{E} = T + \Delta t_\mathcal{E}$ . The $c_k$ value used to calculate the ergodic metric is the average $c_k$ value of all agents on the team. We use the spatial and temporal decompositions to calculate the ergodic metric (\ref{eq:ergodic_metric}). 

We determine the ergodic metric's time sensitivity by differentiating with respect to a fixed time duration $\lambda$ and evaluating it at some fixed time $t$, which results in the costate equation.
        We can additionally show that each agent can generate an independent action that contributes to minimizing the
        entire team's ergodic metric. Let us first define the dynamics of the collective multi-agent system as
        \begin{align} \label{eq:collective_dynamics}
            \dot{x} & = f(x,u) = g(x) + h(x) u \nonumber\\
            & = \begin{bmatrix}
            g_1(x_1) \\
            g_2(x_2) \\
            \vdots \\
            g_N(x_N)
            \end{bmatrix} +
             \begin{bmatrix}
            h_1(x_1) & \ldots & 0\\
            \vdots& \ddots & \\
            0 & & h_N(x_N)
            \end{bmatrix} u.
        \end{align}
        We calculate the adjoint variable of the ergodic objective function as
        \begin{equation}\label{eq:decentralized_adjoint}
            \dot{\rho} = -\mathcal{E}(x(t)) \frac{\partial F_k}{\partial x} - \frac{\partial \Phi}{\partial x} - \frac{\partial f}{\partial x} \rho (t), 
        \end{equation}
where $\Phi$ is a barrier function that is needed to keep the agents within their operating environment. (This barrier function is needed because of the spatial periodicity of Fourier transformations.) $f(x,u)$ represents the dynamics of the multi-agent team. We then calculate
        \begin{equation*}
            \frac{\partial \tilde{F}_k}{\partial x} = \frac{1}{N} \begin{bmatrix}
            \frac{\partial F_k (x_1)}{\partial x_1} \\
            \vdots \\
            \frac{\partial F_k(x_N)}{\partial x_N}
            \end{bmatrix}
            \text{ and }
            \frac{\partial f}{\partial x} =
            \begin{bmatrix}
            \frac{\partial f_1}{\partial x_1} & 0 & \ldots & 0 \\
            0 & \frac{\partial f_{2}}{\partial x_{2}} \\
            \vdots & & \ddots  & \\
            0 &  & & \frac{\partial f_N}{\partial x_N}
            \end{bmatrix},
        \end{equation*}
        where $\frac{\partial f}{\partial x}$ is a block diagonal. 
        As a result, following ~\cite{abraham_decentralized_2018}, we define a controller for the collective swarm that
        minimizes the ergodic metric:
        {\small
        \begin{equation} \label{eq:expanded_control}
            \begin{bmatrix}
            u_{\star,1} (t) \\
            \vdots \\
            u_{\star,_N}(t) \\
            \end{bmatrix}
            =
            -R^{-1}
             \begin{bmatrix}
            h_1(x_1) & \ldots & 0\\
            \vdots& \ddots & \\
            0 & & h_N(x_N)
            \end{bmatrix} ^\top
            \begin{bmatrix}
            \rho_1 (t) \\
            \vdots \\
            \rho_N(t)
            \end{bmatrix}+
             \begin{bmatrix}
             u_{\text{def},1} (t) \\
             \vdots \\
             u_{\text{def},N}(t)
             \end{bmatrix},
        \end{equation}}where $R\in\mathbb{R}^{mN \times mN}$ is a diagonal matrix representing the weights on each swarm member's control inputs and $mN$ is the size of the swarm system control input.
        Since $h(x)$ is block diagonal,  (\ref{eq:expanded_control}) becomes
        \begin{equation}\label{eq:policy_independence}
            u_{\star,j} (t) = -R_j^{-1} h_j(x_j)^T \rho_j(t)
        \end{equation}
        for each agent $j \in \left[ 1, \ldots, N \right]$ and $R_j \in \mathbb{R}^{m \times m}$. Note that the
        $j^\text{th}$ agent does not depend on the $i^\text{th}$ agent. All that is required is that we communicate the
        $c_k$ values between each agent to obtain the time-averaged statistics of the swarm as a whole before computing the control values. Each agent has their own specification of the task target distribution $\phi(x)$ for which they calculate control values. Since each agent calculates its own control values, the computational complexity of our decentralized ergodic control algorithm remains constant with respect to the number of agents in the swarm. Computational complexity will only scale with the dimension of an agent's state (which is not time-varying for these demonstrations).

        If a single agent detaches from the communication network, then that agent will minimize its own ergodic metric. Communication from other agents in the swarm helps the individual agent minimize the energy it expends as it moves through the exploration space since it knows where other agents have recently been from their communicated $c_k$ values. This enables the swarm as a whole to converge to the user's target distribution. The following section describes the local planner we used for obstacle avoidance and low-level control.

        \subsection{Interfacing with RT-RRT$^\star$}\label{sec:rrt}
            The control signal $u(t)$ for each agent is converted to kinematic input $\dot{x}(t)$ where the forward
            simulation $x(t)$ for each individual agent is supplied to the RT-RRT$^\star$ low-level planner as a set of target controls. All safety features, such as rover velocity limits and obstacle avoidance, are handled \emph{after} the user target specification is transformed by the ergodic coverage algorithm. Thus, the controls that are sent to the rover may be different from the output of the ergodic coverage algorithm. This allows the ergodic coverage algorithm to avoid considering obstacle avoidance and
            divides the computational load into two segments: low-level planning for robot control and obstacle
            avoidance, and high-level task planning and adaptation with the ergodic coverage algorithm, which will persistently explore according to the spatial measure defined by the target distribution. Since the ergodic coverage algorithm is temporally driven (i.e., the
            amount of time spent in a region directly impacts future behavior), and because we re-plan at every timestep, any regions the user wants their swarm to explore in that cannot be immediately visited due to obstacles will eventually be visited once the RT-RRT$^\star$ planner calculates a path around the obstacles (assuming an unobstructed path from the agent's current position to these regions exists). 

    \subsection{High-level Planning for Dynamic-Task Adaptation} \label{sec:high_level_planning_dynamic_task_adaptation}

        This section defines how tasks are specified as spatial distributions represented by $\phi(x)$. We also define how new user commands representing new or updated tasks are combined with existing task specifications to enable multimodal distributions to act as descriptions of where agents need to be allocated. We focus on two scenarios: reallocating priority to ``attraction'' regions with high expected information content and lowering priority for ``repulsion'' regions with low expected information content.

        \subsubsection{Specifying Tasks with Easter Eggs (EEs) and Disabling Devices (DDs)}
             We use easter eggs (EE) to denote generic ``attraction'' regions and Disabling Devices (DD) to denote generic ``repulsion'' regions, each of which are Gaussians centered at (latitude, longitude) coordinates in the environment. The positions of EEs and DDs that are specified by the user, or autonomously detected by an agent in the swarm, are sent to each agent in the swarm's network. The
            locations of these elements in the environment are added to the specification $\phi(x)$ by parameterizing the
            expected information content in the environment as a multimodal sum of Gaussians:
            \begin{align}
                \phi(x) &= \frac{1}{\eta} \sum_{i} a_i \exp \left( - \frac{1}{2}\Vert x - x_\text{EE}\Vert^2_{\Sigma^{-1}}\right) \nonumber \\
                    &\ + \frac{1}{\eta} \sum_{j} b_j \left(1-\exp \left( - \frac{1}{2}\Vert x - x_\text{DD}\Vert^2_{\Sigma^{-1}}\right) \right)
            \end{align}
            where $\eta$ is a normalization factor,  $\sum_i a_i = 1$ and $\sum_j b_j = 1$, and $x_\text{EE}, x_\text{DD}$ are the locations of EEs and
            DDs respectively. The parameter $\Sigma$ is the width of the Gaussian region of attraction or repulsion that can be
            tuned based on the size of the environment and the user's desired granularity. We used $\Sigma = \text{diag}(0.01,
            0.01)$ for both EEs and DDs. EEs represent high expected information content regions (attraction regions) while DDs represent
            low expected information content regions (avoidance regions). The resulting distribution is then normalized and
            represented with $10$ Fourier coefficients for each dimension of the exploration space. Exploration space coordinates are
            transformed and scaled to a bounding box of size $[0,1]^2$ for numerical stability.

        \subsubsection{Specifying Tasks Through User Drawings at the Tablet Interfaces}\label{sec:user_command_tablet_interface}
        The interfaces described in Section \ref{sec:tanvas_and_touchscreen} transmit a user's specification (represented as a multimodal distribution drawn by the user on an interface screen) to the team of agents the user is controlling. The user's multimodal task distribution consists of a discretized grid of normalized values that represent the relative importance of different areas of the environment (with respect to each other) that the team of agents operate in. When a user draws an ``attraction'' region on the interface screen, the areas of the interface their finger makes contact with are assigned a value of 1 (the highest level of importance)---this is equivalent to the EE specifications described in the section above, the difference being that user attraction drawings do not have to be Gaussians centered at a particular position of the environment; drawings can take on different shapes that do not resemble Gaussians. Other areas that have not been drawn in are assigned a random noise value between $[0,.001)$ in order
        for the team of agents to generate minimal coverage over the non ``attraction'' areas. When a user draws a ``repulsion'' region, the areas of the screen their finger makes contact with are assigned a value of 0---which is equivalent to the DD specifications described above (with other areas also being assigned random noise between $[0,.001))$. If both ``attraction'' and ''repulsion'' regions are drawn, areas that have not been drawn on are assigned a value of $0.5$, which denotes ``medium'' information content. As above, the resulting distribution of attraction and repulsion regions is normalized and represented using 10 Fourier coefficients for each dimension of the exploration space. 

\section{Small-scale Simulations in ROS}
\label{sec:small_scale_simulations}
\begin{figure}[!htb]
  \centering
  \includegraphics[width=1.0\textwidth]{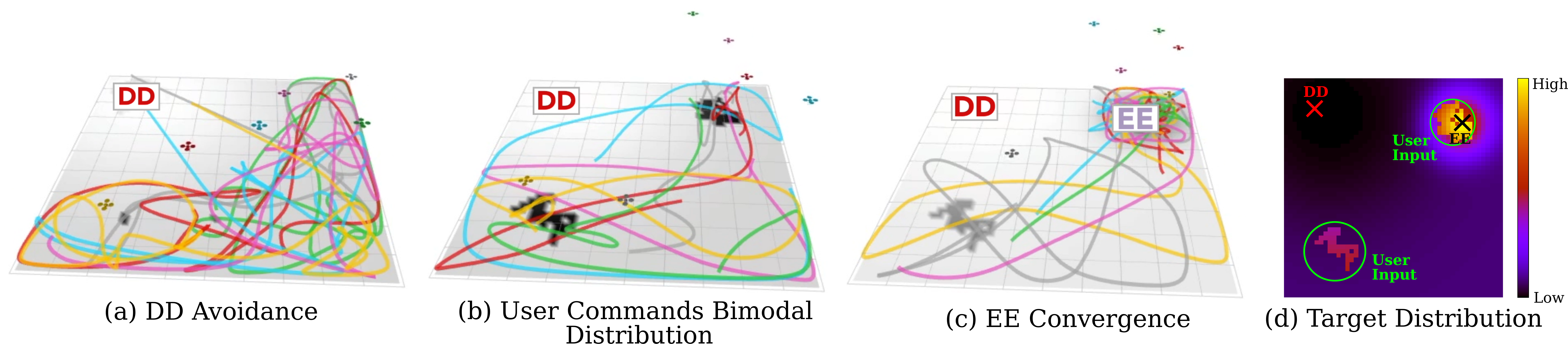}
    \caption{
    \textbf{Simulation of a team of six agents dynamically adapting to the environment while also responding to user commands:} The figures a, b, and c show a time-lapse of the agents' trajectories in response to the different stimuli. (a) The agents discover a DD and explore the rest of the environment while avoiding that location. (b) A user inputs a bimodal target with two areas with high expected information content for the team of agents to explore. The team responds to the user command while continuing to avoid the DD location.(c) The team discovers an EE (in this example the EE has higher expected information content than the user's attraction region drawing input---for other cases, EEs and user attraction regions have an equivalent expected information content) in the environment. The team of agents converges on the EE, while still exploring the user drawing inputs to a lesser extent. The agents continue to avoid the DD location. (d) The target distribution containing the DD, user input, and EE. The ``x'' label in red marks the DD location, while the ``x'' label in black marks the EE location. The circles in green highlight the user target input from the Tanvas interface.   
    }
    \label{fig:combsim}
  \end{figure}

We demonstrate the dynamic response of our system with small-scale proof-of-concept tests in ROS simulations with six agents. We use aerial vehicles for these small-scale tests (as opposed to the ground vehicles used for our simulated demonstrations in Section \ref{sec:stomp_simulations} and field demonstrations in Section \ref{sec:fieldexercises}) because the dynamic response of our system is clearer with faster platforms that do not have to avoid obstacles in the environment. These agents explore an environment and adapt their trajectories to environmental stimuli and user commands sent from a Tanvas interface. Figure \ref{fig:combsim}a shows the team of six agents discover a disabling device (which we use to represent an area of the environment with the lowest possible expected information content) in the upper left hand corner of the environment. 
The agents avoid the disabling device and explore the rest of the environment. The user then sends a bimodal target distribution to the team of agents, represented by the black shaded areas. In Figure \ref{fig:combsim}b, the agents converge to the user's target distribution while they continue to avoid the disabling device. In Figure \ref{fig:combsim}c, the agents discover an easter egg---which we use to represent an area with higher expected information content than the targets sent by the user (the difference in expected information content between EEs and user attraction region drawing inputs are only different for this particular experiment---in all other cases EEs and user drawings that represent attraction regions have the same expected information content values). The team of agents converges on this easter egg but continues to explore near the previous targets sent by the user. The team of agents spends more time near the easter egg due to its higher expected information content. The team of agents also continues to avoid the disabling device. Figure \ref{fig:combsim}d shows the distribution representing the agents' expected information content for the entire environment.

We then simulate a team of six heterogeneous agents exploring and adapting to environmental stimuli based on their individual capabilities. The team of six agents contains five regular agents tasked to explore the environment, while one agent is tasked to remove DDs from the environment. We simulate an agent discovering a DD and communicating its location to its teammates. The regular agents avoid the DD location while the DD blocker agent converges on it to make it safe for any agent that comes close to it.

\begin{figure}[htb]
  \centering
  \includegraphics[width=0.8\textwidth]{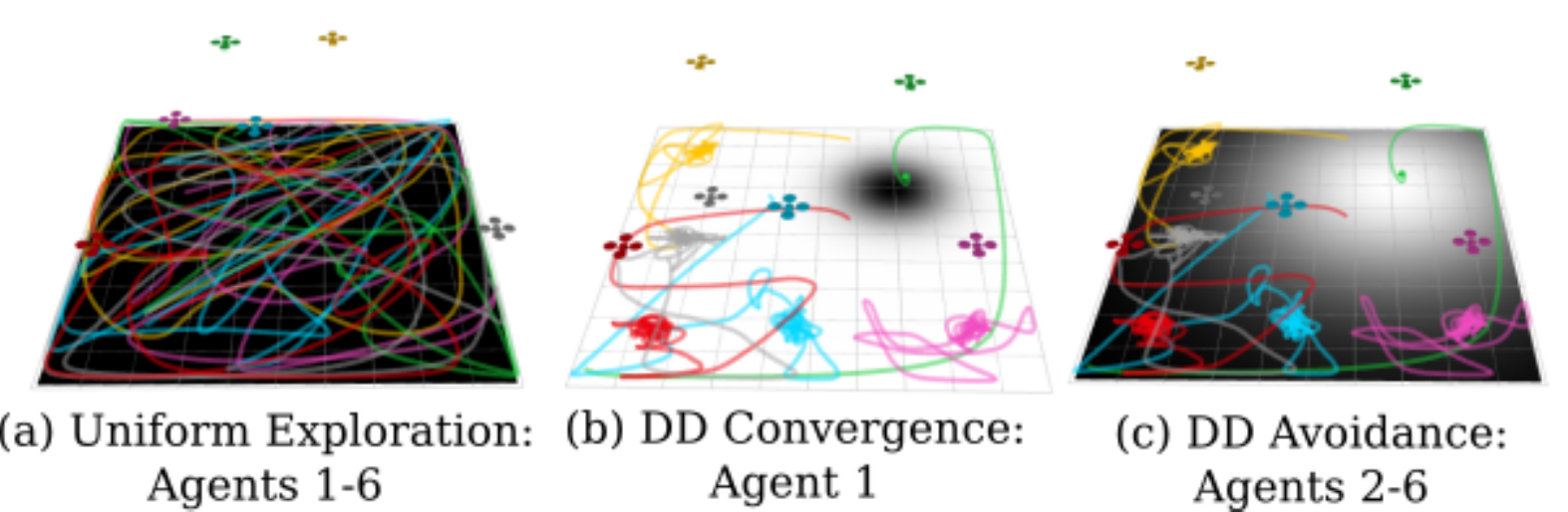}
    \caption{\textbf{Heterogeneous team discovering a disabling device (DD) then dynamically responding to it based on team members' capabilities.} The figures here show a time-lapse of the agent trajectories. The team contains a DD
    blocker agent (agent 1) and non-DD blocking agents (agents 2-6). Thus, agent 1 and agents 2-6 will respond differently to DDs. The team starts by uniformly exploring the environment in (a). Here, the black colored background corresponds to uniform information content in the environment for all of the agents, which leads to uniform coverage. (b) The agents discover a DD in the upper right-hand corner of the environment. The background here shows the information content in the environment from the perspective of the DD blocker agent---the white color corresponds to low information content, while the black color corresponds to high information content (the position of the DD). In this example, information content in the environment denotes where agents in the swarm should position themselves according to their capabilities. The DD blocker agent converges on the DD location (denoted by the black circle), while the other agents scatter and explore areas away from the DD. (c) The information content of the environment from the perspective of the non-DD blocking agents. The non-DD blocking agents avoid the white-colored area of the environment denoting low information content (where the DD is) and explore the black areas of the environment away from the DD.}
    \label{fig:hetsim}
\end{figure}

In Figure \ref{fig:hetsim}a, the team of heterogeneous agents uniformly explores the environment. When a DD location is discovered (Figure \ref{fig:hetsim}b) the DD blocker agent converges on DD to remove it from the environment,  while the rest of the team continues to explore away from the DD (in Figure \ref{fig:hetsim}c). The examples shown in Figures \ref{fig:combsim} and \ref{fig:hetsim} demonstrate our system responding to detected changes in the information content of the environment with homogeneous and heterogeneous teams of six agents. The next section describes how we use STOMP to interface with a larger number of agents and demonstrate the scale-invariance of our system.
  
\begin{figure}[!htb]
    \centering
    \includegraphics[width=0.95\textwidth]{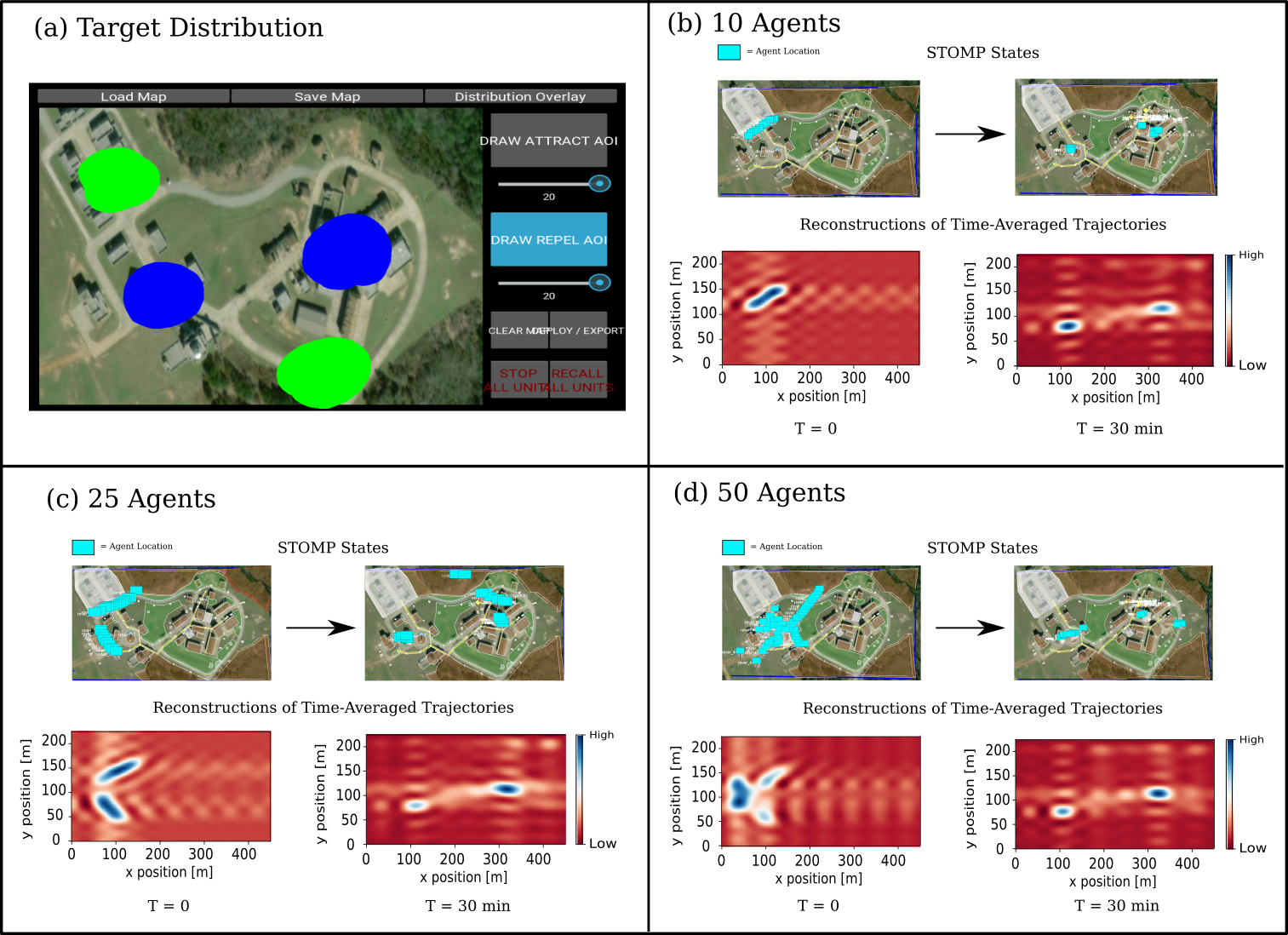}
    \caption{\textbf{Scale Invariance in Target Distribution Convergence with respect to Swarm Size:} Swarms of different sizes will all converge to user target distributions. The user specifies an example target distribution (shown in figure a) which contains two areas with high expected information content (shown in blue) and two areas with low expected information content (shown in green). Figures b, c, and d show swarms containing 10, 25, and 50 agents converging to this user target distribution. The top row of each of these figures b,c,d shows the initial positions of agents in the swarm (denoted by the blue rectangles) at time 0s and their final positions after converging to the user's target distribution at 1800s. The bottom row of each of the figures shows the Fourier reconstruction of the time-average of the agents' trajectories in these swarms at 0s (on the bottom left) and at 1800s (at the bottom right). Despite the different sizes, all three swarms converge to the user target, as seen by the near identical Fourier reconstructions of the swarms' time-averaged trajectories at 1800s.}
    \label{fig:combined_multimodal}
    \vspace{-1em}
\end{figure}

\section{Demonstrations Using STOMP}\label{sec:stomp_simulations}
In this section, we demonstrate scenarios that are relevant to the OFFSET program by using STOMP to interface with up to 50 simulated agents. We demonstrate the high-level benefits our system offers in scenarios that include covering multiple areas of the environment, converging upon a target area and quickly evacuating from it, and simultaneously converging on and surrounding a target area, one of the stated goals for the FX-5 exercises\footnote{We did not perform physical experiments in FX-5 due to the COVID-19 pandemic}. We also show how information collected by the swarm varies according to the size of the swarm, and how our system automatically adapts to an operator's target specification if the swarm loses agents\footnote{The code for our system can be found at: \textbf{https://github.com/MurpheyLab/rover\_decentralized\_ergodic\_control}}.

\subsection{User Specifications Are Scale-Invariant With Respect to Swarm Size}
\label{sec:multimodal}
In swarm exploration scenarios, the operator may want to specify which areas of the environment are more or less likely to contain information. Our system enables an operator to draw an arbitrary, multimodal distribution that contains areas with both high and low expected information content for their swarm to explore. Our decentralized coverage algorithm \cite{abraham_decentralized_2018} plans trajectories for the swarm such that the time agents spend in different areas of the environment is proportional to their expected information content. In the OFFSET setting, this is relevant for operators who seek to use swarms of robots to discover where targets of interest are and which areas should be avoided due to threats. We demonstrate an operator sending a multimodal target distribution to their swarm that contains two areas with high expected information content (which the swarm will spend more time in) and two areas with lower expected information content (which the swarm will spend less time in) on swarms that contain 10, 25, and 50 agents. The results are shown in Figure \ref{fig:combined_multimodal}.

The multimodal target shown in Figure \ref{fig:combined_multimodal} contains two regions with high expected information content (blue) centered on clusters of buildings in the STOMP environment and two areas with low expected information content (green) centered on featureless areas in the environment. The operator monitors the swarm's ``progress'' with respect to the user's target specification through both the normalized ergodic metric plot (which indicates convergence when the steady-state normalized ergodic metric value approaches 0 or plateaus after decreasing for some time) and the plot showing the Fourier reconstruction of the time-average of the agents' trajectories (which provides visual feedback on how closely the Fourier reconstruction of the swarm's time-averaged trajectories matches the operator's target specification). The user knows the swarm has converged to their target specification (and completed the current task) when the Fourier reconstruction plot of the time-average of the agent trajectories in the swarm matches the target specification the user has given through their interface, or when the value of the normalized ergodic metric plot for their system plateaus and remains relatively constant (after decreasing for some time).

\begin{figure}[htb]
    \centering
    \includegraphics[width=0.6\textwidth]{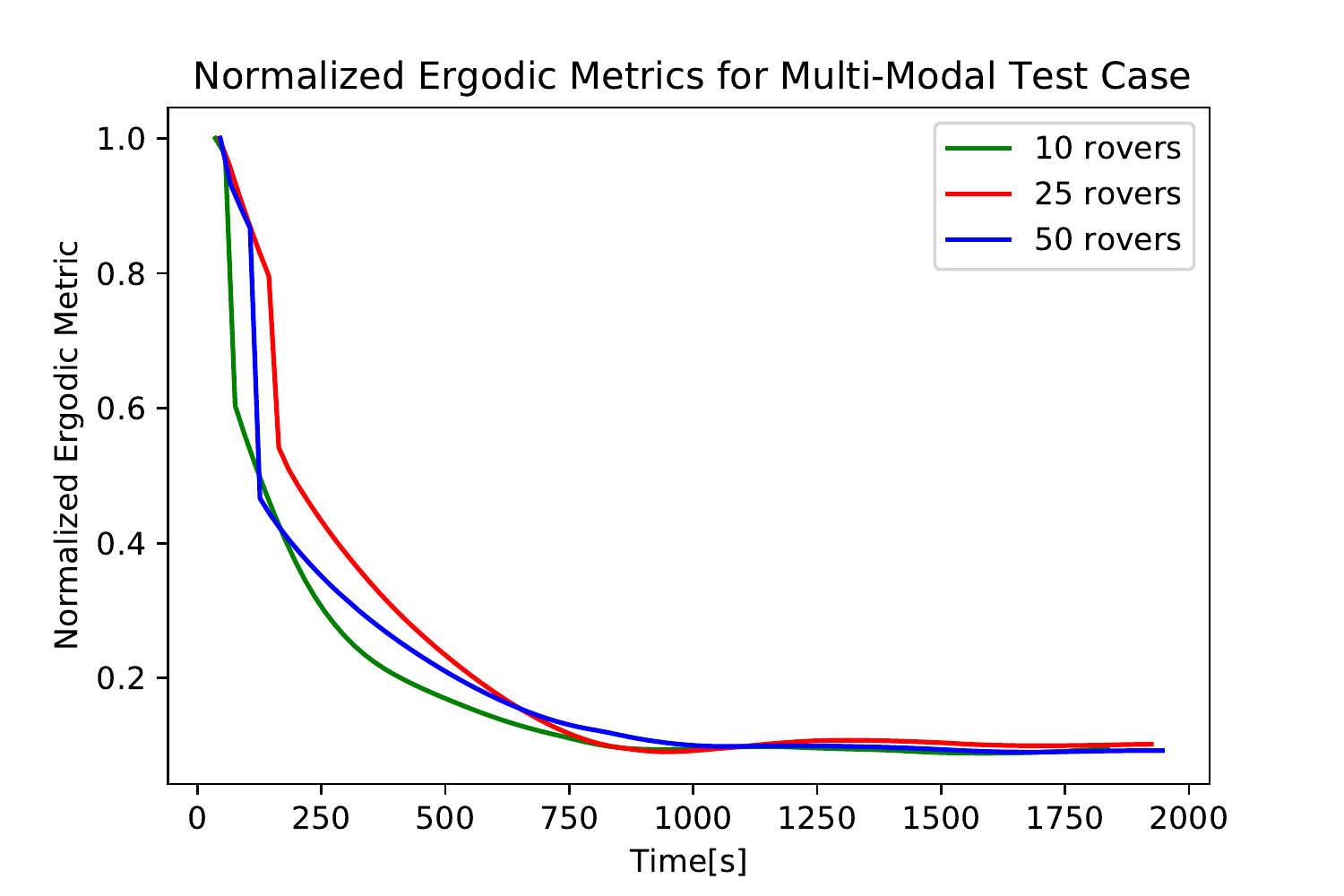}
    \caption{\textbf{Normalized Ergodic Metric Values Converge Across Scale:} Swarms containing 10, 25, and 50 agents converged to the operator's desired multimodal target distribution (given in Figure \ref{fig:combined_multimodal}), as seen through the normalized ergodic metric values over time for all three swarm sizes asymptotically converging to 0. Despite the small difference in the time at which the operator's multimodal target distribution was sent to each differently sized swarm, all three swarms converged to the operator's target after approximately 1200s (the experiments for each swarm size lasted for roughly 1800s in total).}
    \label{fig:multimodal_erg_metric}
\end{figure}

Figure \ref{fig:multimodal_erg_metric} shows a plot of the normalized ergodic metric values over time for three different swarm sizes (10, 25, and 50 agents) running the same multimodal target distribution in the STOMP environment. As seen in Figure \ref{fig:multimodal_erg_metric}, the normalized ergodic metric values of all three swarm sizes converge to near zero at roughly the same time. The Fourier reconstructions of the time-average of the agents' trajectories (shown in the ``Reconstructions of Time-Averaged Trajectories'' plots in Figure \ref{fig:combined_multimodal}) of all three swarms also closely match the operator's original multimodal target specification at the end of each run. Large portions of the agent trajectories for each run are spent in areas of the map with high expected information content. The agent trajectories avoid areas with low expected information content. This indicates that an operator can expect the same system behavior from swarms of different sizes.

\subsection{Dynamically Responding to User Target Re-Specification}\label{sec:converge_repel}
In real-world operations, operators need to be able to quickly re-task their swarm in response to changing environments. We present an example scenario in which a swarm is given an area with high expected information content in the center of the map that then inverts and becomes an area with a low expected information content (an area to avoid). This scenario is relevant to OFFSET in that a rover may discover a disabling device in an area that was previously thought to be ``safe''. The system would have to quickly respond to this new knowledge. It is difficult to achieve this type of behavior with geometric (Voronoi) methods such as \cite{cortes_coverage_2004}, leader-follower methods for dispersion \cite{hsiang_algorithms_2004}, or methods in which a user controls (or selectively takes control of) a single agent \cite{swamy_scaled_2020}. 
\begin{figure}[!htb]
    \centering
    \includegraphics[width=0.7\textwidth]{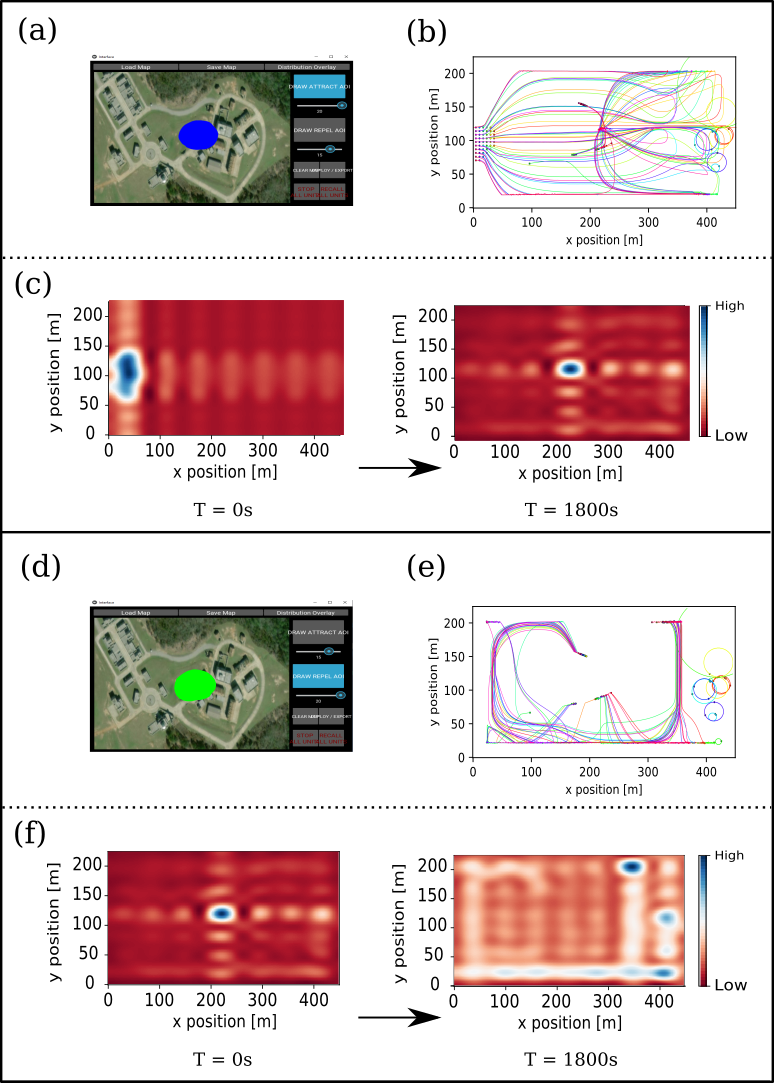}
    \caption{\textbf{Dynamically Responding to User Target Re-Specification:} This figure shows a user specifying their swarm of 50 agents to converge upon a specific area in the center of the environment---the user then re-specifies the center as an area to avoid, which causes their swarm to scatter. (a) The initial user target specification, which is a convergence region (``attraction'' regions are drawn in blue in the interface). (b) Swarm agent trajectories responding to this target---the agents move from their initial positions on the left side of the environment to the center (the different colored lines represent the trajectories of different agents). (c) The Fourier reconstruction of the time-average of these agent trajectories at time t=0s and t=1800s. (d) The user re-specifies their target to be an avoidance region (``repulsion'' regions are drawn in green) which is located at the area that they previously specified to be a convergence region. (e) The agent trajectories moving away from the center of the environment in response to this new target. (f) The Fourier reconstruction of the time-averaged agent trajectories for this new repulsion target. A video of this sequence can be viewed online: \textbf{https://sites.google.com/view/scale-invariant-human-swarm}} 
    \label{fig:converge_and_repel}
\end{figure}
Figure \ref{fig:converge_and_repel} demonstrates this behavior with our system for a team of 50 agents, showing the transition between the team converging on an attraction region and then scattering when the attraction region inverts into a repulsion region. This example is a larger scale version (in terms of number of agents and operating environment) of the example introduced in Section \ref{sec:small_scale_simulations}.

\begin{figure}[!hbt]
    \centering
    \includegraphics[width=0.7\textwidth]{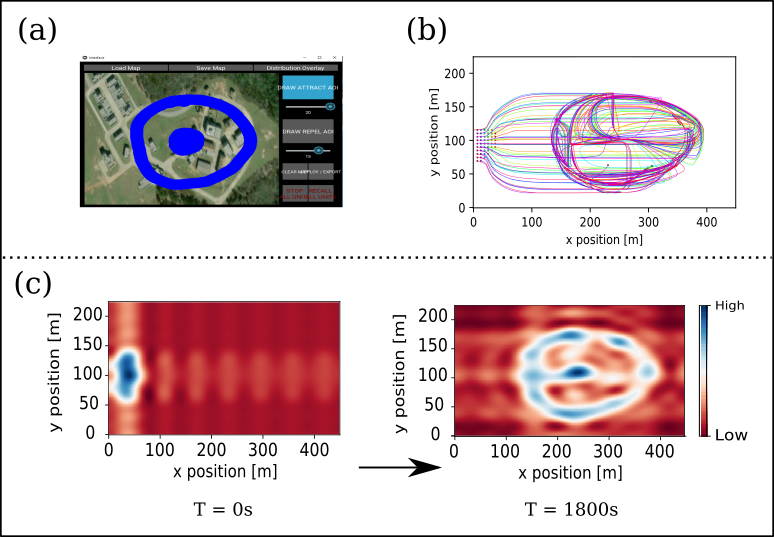}
    \caption{\textbf{Non-trivial Multimodal Demonstrations:} Our system enables swarms to converge to ``non-trivial'' multimodal target distributions. (a) The user's target distribution, which resembles a bullseye. (b) The agent trajectories as the swarm converges to this target (the different colored plot lines represent the trajectories of different agents). (c) The Fourier reconstruction of the time-averaged trajectories of the swarm at t=0s and t=1800s (the time at which the system has converged). As seen in (c), our system automatically allocates a small number of agents to converge on the smaller centroid of the bullseye, and a larger number of agents to converge on the outer ring that surrounds the centroid. Figure b also shows the agents in the smaller centroid and outer ring exchange positions with one other as time tends towards infinity, which indicates that our system is persistent and will constantly work towards converging to the target.}
    \label{fig:bullseye}
\end{figure}
\subsection{Non-trivial Multimodal Distributions (Bullseye)}\label{sec:bullseye}
An advantage of our system is the flexibility with which it provides operators---allowing operators to task swarms to explore areas of any arbitrary shape. Figure \ref{fig:bullseye} shows a swarm of 50 agents converging on a target that resembles a ``bullseye'' with two regions to converge to: a circle surrounded by an outer ring. This target distribution is analogous to a tactical situation in which a swarm must converge on a target while surrounding it, perhaps to prevent the target from escaping \cite{day_responding_2018}. As shown in Figure \ref{fig:bullseye}, our system enables the swarm to successfully converge to this target. Also of note is that the agents will exchange positions with other agents between the center and outer ring of the bullseye as the system converges, which indicates that the swarm automatically adjusts to real-time changes in its time-averaged behavior. Thus the operator does not need to manually reallocate members of the swarm, a task that could be mentally taxing as the total number of agents increases. Instead, the operator can focus on high level strategy and interpreting the information their swarm collects from the environment.

\begin{figure}[h]
    \centering
    \includegraphics[width=0.7\textwidth]{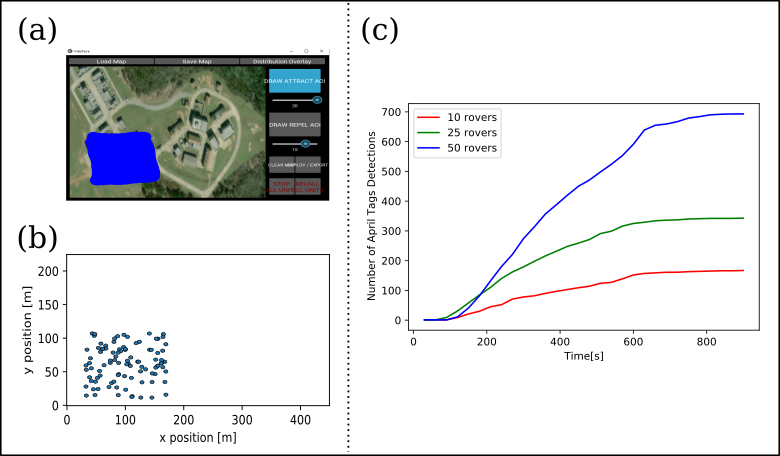}
    \caption{\textbf{Graceful Degradation in the Amount of Information Collected Across Swarms of Different Sizes:} The user sends the same target distribution (a) to swarms of 10, 25, and 50 agents. The user expects there to be information (in the form of April Tags in this example) in the bottom left corner of the environment. (b) The ground truth distribution of April Tags in the environment. As the swarms of different sizes converge to the user's target distributions, they detect the April Tags in the environment. (c) The number of April Tag detections for each swarm after 900s. The swarm of 50 agents detects the most April Tags, followed by the swarms with 25 agents and 10 agents. The April Tag detection curves for all three different swarm sizes have the same shape and trend, which indicates there is a graceful change in information detection across swarm size. Thus, an operator can expect larger swarms to detect more information, but will not have to alter their strategy or mental model with smaller swarms.}
    \label{fig:april_tags}
\end{figure}

\subsection{Graceful Degradation in the Amount of Information Collected Across Swarms of Different Sizes}\label{sec:april_tags}
We also demonstrate the graceful degradation of our system---our system continuing to operate under unexpected agent loss and enabling the operator to use the same strategy or mental model with swarms of different sizes---with respect to the amount of information the swarm collects. For ``information'' we use simulated April Tags that represent different entities relevant to the OFFSET program (IED, High Value Target, etc.). 
We randomly distribute these April Tags throughout the STOMP simulation environment. As shown in Figure \ref{fig:april_tags}, the operator draws a target distribution that uniformly covers an area of the map in which they expect their swarm to obtain information. As the figure shows, the largest swarm detects the highest number of April Tags. However, there was a graceful degradation in the amount of information collected for the smaller swarms, indicating that our system is suitable for uncovering information at any scale, and that an operator can use our system with different numbers of agents knowing that the amount of information collected will gracefully increase or decrease if new agents are added to the swarm or if agents are disabled during operation.

\begin{figure}[!htb]
    \centering
    \includegraphics[width=0.95\textwidth]{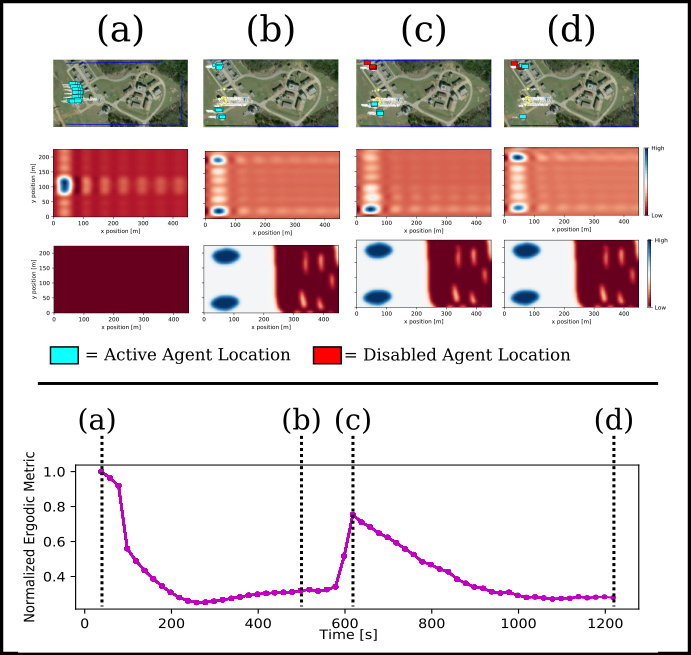}
    \caption{\textbf{Recovery from Agent Failure:} In column (a) of the top half of this figure, the operator starts by specifying a multimodal distribution with two areas containing high expected information content in the top and bottom left corners of the environment. Their swarm of 25 agents converges to this target distribution, shown in column (b). In column (c), an EMP (electromagnetic pulse) device disables all of the agents in the top left corner of the environment. The swarm automatically adapts to the drop in the number of accessible agents by reallocating agents from the bottom left corner of the environment to the top right corner. The system then successfully converges to the original target, as seen in column (d). The bottom row of the figure shows the  normalized ergodic metric plot for the swarm over the course of this experiment. The normalized ergodic metric starts at 1.0 at the timestamp shown in column (a). The normalized ergodic metric smoothly decreases to the value corresponding to column (b) as the system converges to the user target distribution. After the EMP occurs, there is a spike in the normalized ergodic metric plot (column (c)). The system is now ``far'' from converging because all of the agents in one of the target areas have been disabled. However, the system is not as far as it was from converging at the start of the experiment in column (a) since some of the agents are still located at one of the user target areas. After the swarm reallocates members to the top left corner, the normalized ergodic metric converges to near zero again, shown in column (d).}
    \label{fig:kill_agents}
\end{figure}
\subsection{Scale-Invariant Continuity in the Event of Agent Failure}\label{sec:kill_agents}
The scale-invariant continuity of our system also enables operators to retain the same strategy for executing a task even if agents in their swarm unexpectedly fail. Figure \ref{fig:kill_agents} shows the results of an operator running a multimodal target distribution with 50 agents, and then losing 25 of them to attrition (due to a simulated EMP) as they converge towards the target. The figure shows a spike in the normalized ergodic metric, which indicates that the swarm, now containing less agents, is further from converging to the target distribution than the original team was at the instant before 25 of the agents were disabled. Despite this, the system is able to continue converging towards the original target without user intervention, which indicates our system can ``recover'' from the sudden loss of agents without methods such as network topology reconfiguration \cite{chen_toward_2020}. Thus, an operator does not have to abandon their current specification (or mental model of how to complete a task) if they lose agents. The remaining agents will converge to the existing target given enough time. 

\section{Field Demonstration}
\label{sec:fieldexercises}
We demonstrated the above operator interface at the FX-3 OFFSET field test in December 2019. The purpose of this demonstration was to enable operators to control the robots from the tablet interface (though we mention here that some of the testing included commanding the robots from the command line by directly specifying distributions).  We focused on the case of identifying a particular target area and then responding to a dangerous event that might disable other robots. 

\begin{figure}[htb]
  \centering
  \includegraphics[width=0.55\textwidth]{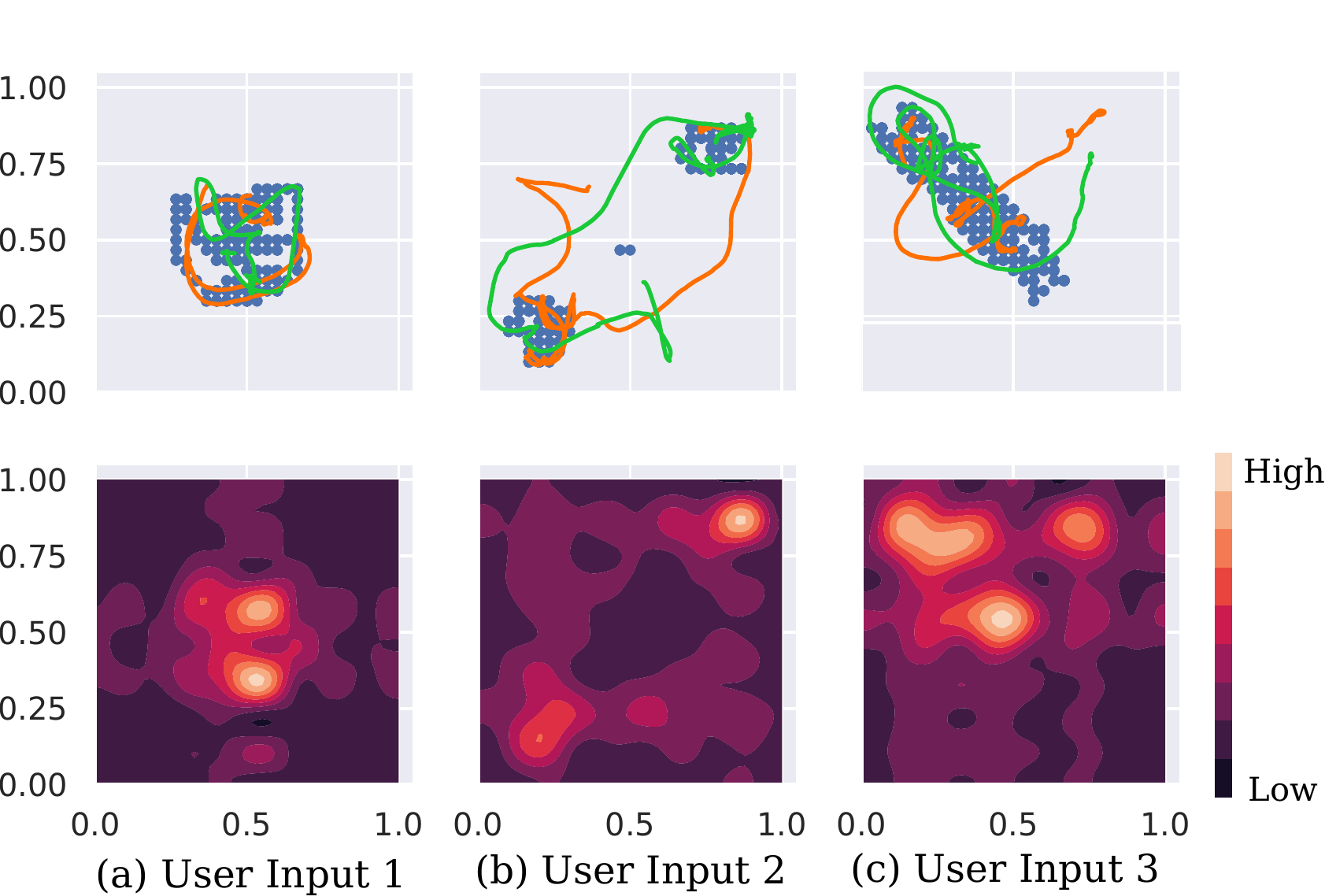}
    \caption{\textbf{Real-world agents dynamically respond to the user's target distribution sent through the Tanvas:} The figures at the top show the time-lapse of the real world agents' trajectories (shown in orange and green). The bottom figures show the Fourier reconstruction of the time-average of these trajectories. When the user updates their target distribution (represented in sub-figures a,b,c with the blue dots) the robots update their trajectories to converge to this new target.
    }
  \label{fig:tanvas_timelapse}
  \vspace{-1em}
\end{figure}

In Figure \ref{fig:tanvas_timelapse}, several real world rovers adapt to the operator's specifications, which are represented by the dotted blue areas on the top row of Figure \ref{fig:tanvas_timelapse}. As the real world rovers receive new inputs through the Tanvas tablet interface, they update their target distributions and spend more time in areas of the environment with higher expected information density. For this real-world demonstration, we started with four real-world rovers which then dropped down to two due to hardware issues. Despite the hardware issues, our system was able to adapt to the user target specifications with the drop in team size. The remaining agents updated their trajectories to compensate for the loss of their two team members.
\quad

Next, we demonstrate a team of four real-world rovers exploring their environment and dynamically adapting to environmental stimuli they encounter in real-time. Through the command-line, we simulate one of the rovers discovering a high value target and then sending the location of this high value target to its teammates. Each member of the team updates its own target specification. We then transform the high value target location into a disabling device, which causes the team members to scatter to avoid the area.

\begin{figure}[htb]
  \centering
  \includegraphics[width=\textwidth]{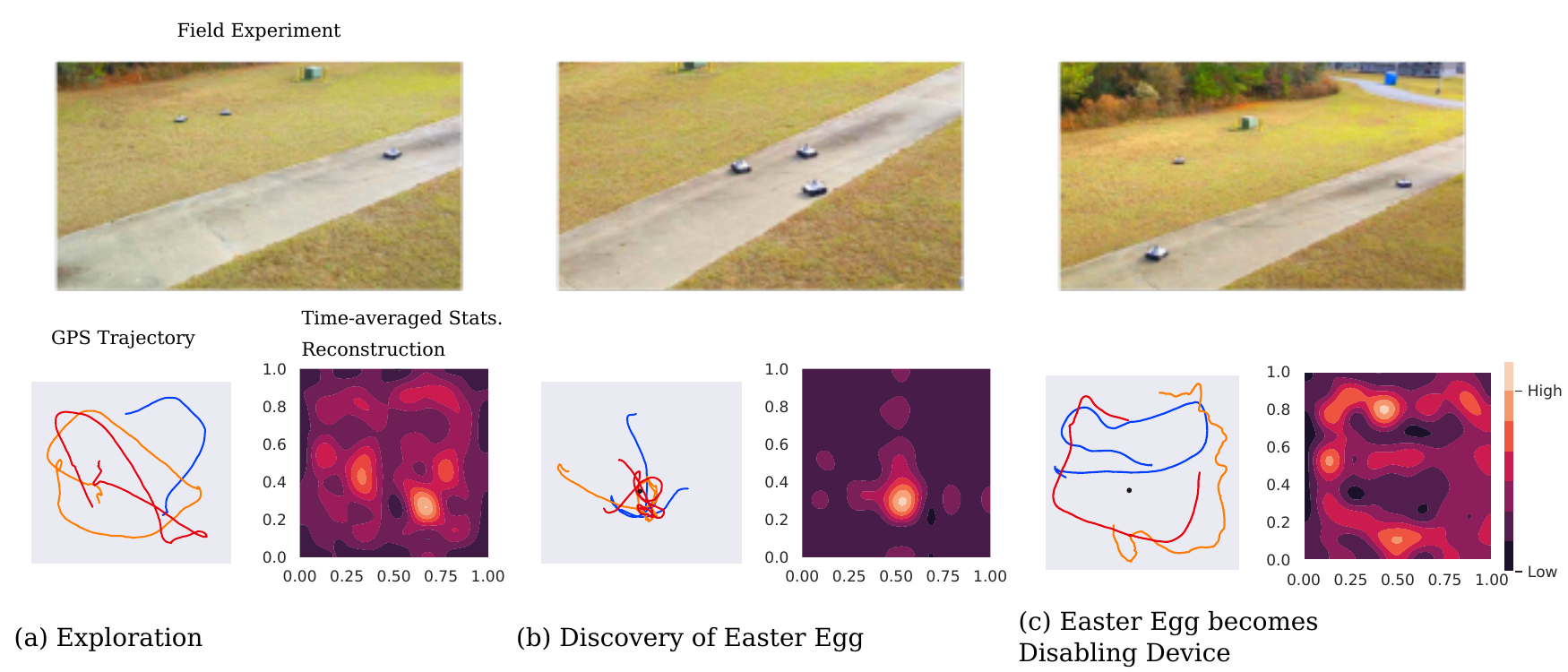}
    \caption{\textbf{Inverting a User Specification Produces the Same Behavior in the Real-World as it does in Simulation:} The real-world rovers converge on a target with high expected information content. This target then transforms into a disabling device that the rovers must avoid. The top figure shows pictures of the real-world testing environment. In each bottom figure, the left shows the robots' trajectories (in red, orange, and blue) while the right shows the Fourier reconstructions of the time-average of these trajectories. In Figure (a), the real-world rovers start by uniformly covering their environment. We then add an area with high expected information content (which could represent a high value target) via the command line to the center of the environment. The rovers update their target distribution with this high value target location (shown in Figure b), and converge towards the center. We then change the high value target to a DD. Figure c shows the rover trajectories and the Fourier reconstruction of the time-average of these trajectories after 40 seconds --- which was the approximate time the rovers needed to move away from the location.}
    \label{ied_timelapse}
\end{figure}

Figure \ref{ied_timelapse} shows the results at each stage of the scenario described above. The rovers start out uniformly exploring the task space (Figure~\ref{ied_timelapse}a) until a high value target is discovered (Figure~\ref{ied_timelapse}b). The rover that discovered the high value target shares the target's location to its teammates. Each rover in the team updates its target distribution in a decentralized manner. Note that when the rovers converge to the target location they do not move to the peak and wait --- instead, they generate persistent, exploratory trajectories over the environment (see the bottom row that contains the Fourier reconstruction of the time-average of the rover trajectories (Figure~\ref{ied_timelapse}). We then transform the high value target into a disabling device, which causes the rovers to scatter (see Figure~\ref{ied_timelapse}c). As the rovers scatter, they uniformly explore areas of the environment that are away from the disabling device. These examples show that our system works in the real-world, in real-time, with an operator in the loop.

\section{Discussion, Lessons Learned, and Future Work}\label{sec:lessonslearned}
In this work, we developed a scale-invariant interface framework for real-time human-swarm control. Our touchscreen interface enables users to specify behavior for a swarm in a setting that is scale-invariant---the user specification does not depend on the number of agents---persistent---the specification remains active until the user specifies a new command---and real-time---the user can interact with and interrupt the swarm at any time.  Our decentralized ergodic coverage algorithm enables the swarm to converge to user-specified targets, including in the setting of a large percentage of the agents unexpectedly failing. We demonstrated our system with simulated agents using command software developed by Raytheon BBN for the DARPA OFFSET program with up to 50 agents, and also demonstrated our system working in the field, in real-time, with a human operator in the loop for several scenarios. 

We note that using STOMP to interface with simulated agents requires substantial computation, since the simulation is modeling every agent and their on-board computing.  This practically makes simulation for hundreds of agents challenging; although this computational limitation is not present for a real physical swarm, where each agent is equipped with its own computational resource, it means that mission planning (which inevitably involves some simulation) is dependent on significant computational power.  Future work includes distributing nodes across multiple computers and using more, and more powerful, CPUs to enable simulation of hundreds/thousands of agents.  

Field tests in the OFFSET FX-3 program revealed that the algorithms we present here were capable of persistent coverage even in the presence of long duration network loss.  This practical result that was evident in the experiments themselves is not an obvious outcome for our algorithms; indeed, it is not clear why ergodic coverage may be insensitive to long duration network dropouts.  The authors' basic intuition is that coverage goals, specified by a particular distribution (either specified by a user or determined autonomously) have slower time constants than trajectory specifications.  As a result, even though the trajectories may be executed at high speed, the rate at which the distribution representing coverage changes is at a much lower speed, leading to an overall low sensitivity to network failures.  Future work includes analytically justifying this---including counter-example cases where this intuition fails.  

In other future work, we are interested in creating a swarm that has closed-loop capability: automatically updating its target distribution based upon the environmental stimuli each agent receives (as opposed to simulating this capability by manually adding areas containing high and low expected information content to the environment). Doing so would further reduce the burden on a user.  We also want to investigate how to balance these automatic target distribution updates with updates made by the human operator. Furthermore, we would like to perform field tests that involve dynamic adversarial agents, such as persons wearing April Tags moving throughout the operating environment. (Such tests had originally been within the scope of the OFFSET field tests, but practical limitations and disruption due to the COVID-19 pandemic made them experimentally impractical.) 

\section{Acknowledgements}\label{sec:acknowledgements}
This material is based on work supported by the Defense Advanced Research Projects Agency (DARPA) project under a DARPA project (funded through Space and Naval Warfare Systems Center Pacific grant N660011924024) and the OFFSET SPRINT grant HR00112020035, and by the National Science Foundation under under grant CNS 1837515. The views, opinions and/or findings expressed are those of the author and should not be interpreted as representing the official views or policies of the Department of Defense or the U.S. Government.
Author Joel Meyer was supported by a National Defense Science and Engineering Graduate Fellowship. Authors Millicent Schlafly and Katarina Popovic were supported by National Science Foundation Graduate Research Fellowships.

\bibliographystyle{apalike}
\bibliography{jfr2022_refs}

\begin{thebibliography}{}

\bibitem[Abraham and Murphey, 2018]{abraham_decentralized_2018}
Abraham, I. and Murphey, T.~D. (2018).
\newblock Decentralized {Ergodic} {Control}: {Distribution}-{Driven} {Sensing}
  and {Exploration} for {Multiagent} {Systems}.
\newblock {\em IEEE Robotics and Automation Letters}, 3(4).

\bibitem[Alonso-Mora et~al., 2019]{alonso-mora_distributed_2019}
Alonso-Mora, J., Montijano, E., Nägeli, T., Hilliges, O., Schwager, M., and
  Rus, D. (2019).
\newblock Distributed multi-robot formation control in dynamic environments.
\newblock {\em Autonomous Robots}, 43(5):1079--1100.

\bibitem[Ayanian et~al., 2014]{ayanian_controlling_2014}
Ayanian, N., Spielberg, A., Arbesfeld, M., Strauss, J., and Rus, D. (2014).
\newblock Controlling a team of robots with a single input.
\newblock In {\em 2014 {IEEE} {International} {Conference} on {Robotics} and
  {Automation} ({ICRA})}.

\bibitem[Balch and Arkin, 1998]{balch_behavior-based_1998}
Balch, T. and Arkin, R. (1998).
\newblock Behavior-based formation control for multirobot teams.
\newblock {\em IEEE Transactions on Robotics and Automation}, 14(6):926--939.

\bibitem[Baxter et~al., 2007]{baxter_multi-robot_2007}
Baxter, J.~L., Burke, E.~K., Garibaldi, J.~M., and Norman, M. (2007).
\newblock Multi-{Robot} {Search} and {Rescue}: {A} {Potential} {Field} {Based}
  {Approach}.
\newblock In {\em Autonomous {Robots} and {Agents}}, pages 9--16. Springer.

\bibitem[Bevacqua et~al., 2015]{bevacqua_mixed-initiative_2015}
Bevacqua, G., Cacace, J., Finzi, A., and Lippiello, V. (2015).
\newblock Mixed-{Initiative} {Planning} and {Execution} for {Multiple} {Drones}
  in {Search} and {Rescue} {Missions}.
\newblock In {\em 2015 {International} {Conference} on {Automated} {Planning}
  and {Scheduling}}.

\bibitem[Chen et~al., 2020]{chen_toward_2020}
Chen, W., Liu, J., Guo, H., and Kato, N. (2020).
\newblock Toward {Robust} and {Intelligent} {Drone} {Swarm}: {Challenges} and
  {Future} {Directions}.
\newblock {\em IEEE Network}, 34(4):278--283.

\bibitem[Cortes et~al., 2006]{cortes_robust_2006}
Cortes, J., Martinez, S., and Bullo, F. (2006).
\newblock Robust rendezvous for mobile autonomous agents via proximity graphs
  in arbitrary dimensions.
\newblock {\em IEEE Transactions on Automatic Control}, 51(8):1289--1298.

\bibitem[Cortes et~al., 2004]{cortes_coverage_2004}
Cortes, J., Martinez, S., Karatas, T., and Bullo, F. (2004).
\newblock Coverage control for mobile sensing networks.
\newblock {\em IEEE Transactions on Robotics and Automation}, 20(2):243--255.

\bibitem[Day et~al., 2018]{day_responding_2018}
Day, M., Strickland, L., Squires, E., DeMarco, K., and Pippin, C. (2018).
\newblock Responding to unmanned aerial swarm saturation attacks with
  autonomous counter-swarms.
\newblock In {\em Ground/{Air} {Multisensor} {Interoperability}, {Integration},
  and {Networking} for {Persistent} {ISR} {IX}}.

\bibitem[Diaz-Mercado et~al., 2015]{diaz-mercado_distributed_2015}
Diaz-Mercado, Y., Lee, S.~G., and Egerstedt, M. (2015).
\newblock Distributed dynamic density coverage for human-swarm interactions.
\newblock In {\em 2015 {American} {Control} {Conference} ({ACC})}.

\bibitem[Durantin et~al., 2014]{durantin_using_2014}
Durantin, G., Gagnon, J.-F., Tremblay, S., and Dehais, F. (2014).
\newblock Using near infrared spectroscopy and heart rate variability to detect
  mental overload.
\newblock {\em Behavioural Brain Research}, 259:16--23.

\bibitem[Gateau et~al., 2016]{gateau_considering_2016}
Gateau, T., Chanel, C. P.~C., Le, M.-H., and Dehais, F. (2016).
\newblock Considering human's non-deterministic behavior and his availability
  state when designing a collaborative human-robots system.
\newblock In {\em 2016 {IEEE}/{RSJ} {International} {Conference} on
  {Intelligent} {Robots} and {Systems} ({IROS})}.

\bibitem[Ghaffarkhah et~al., 2011]{ghaffarkhah_dynamic_2011}
Ghaffarkhah, A., Yan, Y., and Mostofi, Y. (2011).
\newblock Dynamic coverage of time-varying environments using a mobile
  robot---{A} communication-aware perspective.
\newblock In {\em 2011 {IEEE} {GLOBECOM} {Workshops}}.

\bibitem[Giles and Giammarco, 2017]{giles_mission-based_2017}
Giles, K. and Giammarco, K. (2017).
\newblock Mission-based {Architecture} for {Swarm} {Composability} ({MASC}).
\newblock {\em Procedia Computer Science}, 114:57--64.

\bibitem[Goodrich et~al., 2012]{goodrich_leadership_2012}
Goodrich, M.~A., Kerman, S., and Jung, S.-Y. (2012).
\newblock On {Leadership} and {Inﬂuence} in {Human}-{Swarm} {Interaction}.
\newblock In {\em {AAAI} {Fall} {Symposium}: {Human} {Control} of {Bioinspired}
  {Swarms}}.

\bibitem[Hsiang et~al., 2004]{hsiang_algorithms_2004}
Hsiang, T.-R., Arkin, E.~M., Bender, M., Fekete, S.~P., and Mitchell, J. S.~B.
  (2004).
\newblock {\em Algorithms for {Rapidly} {Dispersing} {Robot} {Swarms} in
  {Unknown} {Environments}}, volume~V of {\em Algorithmic {Foundations} of
  {Robotics}}.
\newblock Springer.

\bibitem[Hsieh et~al., 2007]{hsieh_adaptive_2007}
Hsieh, M.~A., Cowley, A., Keller, J.~F., Chaimowicz, L., Grocholsky, B., Kumar,
  V., Taylor, C.~J., Endo, Y., Arkin, R.~C., Jung, B., Wolf, D.~F., Sukhatme,
  G.~S., and MacKenzie, D.~C. (2007).
\newblock Adaptive teams of autonomous aerial and ground robots for situational
  awareness.
\newblock {\em Journal of Field Robotics}, 24(11-12):991--1014.

\bibitem[Kato et~al., 2009]{kato_multi-touch_2009}
Kato, J., Sakamoto, D., Inami, M., and Igarashi, T. (2009).
\newblock Multi-touch interface for controlling multiple mobile robots.
\newblock In {\em {CHI} '09 {Extended} {Abstracts} on {Human} {Factors} in
  {Computing} {Systems}}.

\bibitem[Kim et~al., 2020]{kim_user-defined_2020}
Kim, L.~H., Drew, D.~S., Domova, V., and Follmer, S. (2020).
\newblock User-defined {Swarm} {Robot} {Control}.
\newblock In {\em 2020 {CHI} {Conference} on {Human} {Factors} in {Computing}
  {Systems}}.

\bibitem[Kolling et~al., 2012]{kolling_towards_2012}
Kolling, A., Nunnally, S., and Lewis, M. (2012).
\newblock Towards human control of robot swarms.
\newblock In {\em 2012 {ACM}/{IEEE} {International} {Conference} on
  {Human}-{Robot} {Interaction}}.

\bibitem[Lee et~al., 2011]{lee_haptic_2011}
Lee, D., Franchi, A., Giordano, P.~R., Son, H.~I., and Bülthoff, H.~H. (2011).
\newblock Haptic teleoperation of multiple unmanned aerial vehicles over the
  internet.
\newblock In {\em 2011 {IEEE} {International} {Conference} on {Robotics} and
  {Automation}}.

\bibitem[Lee et~al., 2010]{lee_tracking_2010}
Lee, G., Chong, N.~Y., and Christensen, H. (2010).
\newblock Tracking multiple moving targets with swarms of mobile robots.
\newblock {\em Intelligent Service Robotics}, 3(2):61--72.

\bibitem[Leonard et~al., 2010]{leonard_coordinated_2010}
Leonard, N.~E., Paley, D.~A., Davis, R.~E., Fratantoni, D.~M., Lekien, F., and
  Zhang, F. (2010).
\newblock Coordinated control of an underwater glider fleet in an adaptive
  ocean sampling field experiment in {Monterey} {Bay}.
\newblock {\em Journal of Field Robotics}, 27(6):718--740.

\bibitem[Mathew and Mezić, 2011]{mathew_metrics_2011}
Mathew, G. and Mezić, I. (2011).
\newblock Metrics for ergodicity and design of ergodic dynamics for multi-agent
  systems.
\newblock {\em Physica D: Nonlinear Phenomena}, 240(4-5):432--442.

\bibitem[McCammon et~al., 2021]{mccammon_ocean_2021}
McCammon, S., Marcon~dos Santos, G., Frantz, M., Welch, T.~P., Best, G.,
  Shearman, R.~K., Nash, J.~D., Barth, J.~A., Adams, J.~A., and Hollinger,
  G.~A. (2021).
\newblock Ocean front detection and tracking using a team of heterogeneous
  marine vehicles.
\newblock {\em Journal of Field Robotics}, 38(6):854--881.

\bibitem[Micire et~al., 2009]{micire_analysis_2009}
Micire, M., Desai, M., Courtemanche, A., Tsui, K.~M., and Yanco, H.~A. (2009).
\newblock Analysis of natural gestures for controlling robot teams on
  multi-touch tabletop surfaces.
\newblock In {\em 2009 {ACM} {International} {Conference} on {Interactive}
  {Tabletops} and {Surface}}.

\bibitem[Miller et~al., 2016]{miller_ergodic_2016}
Miller, L.~M., Silverman, Y., MacIver, M.~A., and Murphey, T.~D. (2016).
\newblock Ergodic exploration of distributed information.
\newblock {\em IEEE Transactions on Robotics}, 32(1):36--52.

\bibitem[Naderi et~al., 2015]{naderi_rrt}
Naderi, K., Rajamäki, J., and Hämäläinen, P. (2015).
\newblock {RT-RRT*}: a real-time path planning algorithm based on {RRT*}.
\newblock In {\em {Proceedings} of the {8th} {ACM} {SIGGRAPH} {Conference} on
  {Motion} in {Games}}.

\bibitem[Nagi et~al., 2014]{nagi_human-swarm_2014}
Nagi, J., Giusti, A., Gambardella, L.~M., and Di~Caro, G.~A. (2014).
\newblock Human-swarm interaction using spatial gestures.
\newblock In {\em 2014 {IEEE}/{RSJ} {International} {Conference} on
  {Intelligent} {Robots} and {Systems}}.

\bibitem[Ogren et~al., 2004]{ogren_cooperative_2004}
Ogren, P., Fiorelli, E., and Leonard, N. (2004).
\newblock Cooperative control of mobile sensor networks: {Adaptive} gradient
  climbing in a distributed environment.
\newblock {\em IEEE Transactions on Automatic Control}, 49(8):1292--1302.

\bibitem[Olley et~al., 2020]{olley_electronic_2020}
Olley, M. F.~D., Peshkin, M.~A., and Colgate, J.~E. (2020).
\newblock Electronic {Controller} {Haptic} {Display} with {Simultaneous}
  {Sensing} and {Actuation}.

\bibitem[Pimenta et~al., 2008]{pimenta_sensing_2008}
Pimenta, L. C.~A., Kumar, V., Mesquita, R.~C., and Pereira, G. A.~S. (2008).
\newblock Sensing and coverage for a network of heterogeneous robots.
\newblock In {\em 2008 47th {IEEE} {Conference} on {Decision} and {Control}}.

\bibitem[Pinciroli and Beltrame, 2016]{pinciroli_buzz_2016}
Pinciroli, C. and Beltrame, G. (2016).
\newblock Buzz: {An} extensible programming language for heterogeneous swarm
  robotics.
\newblock In {\em 2016 {IEEE}/{RSJ} {International} {Conference} on
  {Intelligent} {Robots} and {Systems} ({IROS})}.

\bibitem[Podevijn et~al., 2014]{podevijn_gesturing_2014}
Podevijn, G., O’Grady, R., Nashed, Y. S.~G., and Dorigo, M. (2014).
\newblock Gesturing at {Subswarms}: {Towards} {Direct} {Human} {Control} of
  {Robot} {Swarms}.
\newblock In {\em 2014 {Towards} {Autonomous} {Robotic} {Systems} {Conference}
  ({TAROS})}.

\bibitem[Prabhakar et~al., 2020]{prabhakar_ergodic_2020}
Prabhakar, A., Abraham, I., Taylor, A., Schlafly, M., Popovic, K., Diniz, G.,
  Teich, B., Simidchieva, B., Clark, S., and Murphey, T. (2020).
\newblock Ergodic {Specifications} for {Flexible} {Swarm} {Control}: {From}
  {User} {Commands} to {Persistent} {Adaptation}.
\newblock In {\em Robotics: {Science} and {Systems} 2020}.

\bibitem[Sampedro et~al., 2016]{sampedro_flexible_2016}
Sampedro, C., Bavle, H., Sanchez-Lopez, J.~L., Fernández, R. A.~S.,
  Rodríguez-Ramos, A., Molina, M., and Campoy, P. (2016).
\newblock A flexible and dynamic mission planning architecture for {UAV} swarm
  coordination.
\newblock In {\em 2016 {International} {Conference} on {Unmanned} {Aircraft}
  {Systems} ({ICUAS})}.

\bibitem[Schwager et~al., 2009]{schwager_decentralized_2009}
Schwager, M., Rus, D., and Slotine, J.-J. (2009).
\newblock Decentralized, {Adaptive} {Coverage} {Control} for {Networked}
  {Robots}.
\newblock {\em The International Journal of Robotics Research}, 28(3):357--375.

\bibitem[Setter et~al., 2015]{setter_team-level_2015}
Setter, T., Kawashima, H., and Egerstedt, M. (2015).
\newblock Team-level properties for haptic human-swarm interactions.
\newblock In {\em 2015 {American} {Control} {Conference} ({ACC})}.

\bibitem[Smith et~al., 2011]{smith_persistent_2011}
Smith, R.~N., Schwager, M., Smith, S.~L., Jones, B.~H., Rus, D., and Sukhatme,
  G.~S. (2011).
\newblock Persistent ocean monitoring with underwater gliders: {Adapting}
  sampling resolution.
\newblock {\em Journal of Field Robotics}, 28(5):714--741.

\bibitem[Smith et~al., 2012]{smith_persistent_2012}
Smith, S.~L., Schwager, M., and Rus, D. (2012).
\newblock Persistent {Robotic} {Tasks}: {Monitoring} and {Sweeping} in
  {Changing} {Environments}.
\newblock {\em IEEE Transactions on Robotics}, 28(2):410--426.

\bibitem[Song and Kumar, 2002]{song_potential_2002}
Song, P. and Kumar, V. (2002).
\newblock A potential field based approach to multi-robot manipulation.
\newblock In {\em 2002 {IEEE} {International} {Conference} on {Robotics} and
  {Automation}}.

\bibitem[Swamy et~al., 2020]{swamy_scaled_2020}
Swamy, G., Reddy, S., Levine, S., and Dragan, A.~D. (2020).
\newblock Scaled {Autonomy}: {Enabling} {Human} {Operators} to {Control}
  {Robot} {Fleets}.
\newblock In {\em 2020 {IEEE} {International} {Conference} on {Robotics} and
  {Automation} ({ICRA})}.

\bibitem[Tsykunov et~al., 2019]{tsykunov_swarmtouch_2019}
Tsykunov, E., Agishev, R., Ibrahimov, R., Labazanova, L., Tleugazy, A., and
  Tsetserukou, D. (2019).
\newblock {SwarmTouch}: {Guiding} a {Swarm} of {Micro}-{Quadrotors} {With}
  {Impedance} {Control} {Using} a {Wearable} {Tactile} {Interface}.
\newblock {\em IEEE Transactions on Haptics}, 12(3):363--374.

\bibitem[Walker et~al., 2013]{walker_human_2013}
Walker, P., Amraii, S.~A., Lewis, M., Chakraborty, N., and Sycara, K. (2013).
\newblock Human {Control} of {Leader}-{Based} {Swarms}.
\newblock In {\em 2013 {IEEE} {International} {Conference} on {Systems}, {Man},
  and {Cybernetics}}.

\bibitem[Wei et~al., 2013]{wei_agent-based_2013}
Wei, Y., Madey, G.~R., and Blake, M.~B. (2013).
\newblock Agent-based {Simulation} for {UAV} {Swarm} {Mission} {Planning} and
  {Execution}.
\newblock In {\em 2013 {Agent}-{Directed} {Simulation} {Symposium}}.

\end{thebibliography}






\end{document}